\documentclass[sigconf]{acmart}

\usepackage{times}
\usepackage{latexsym}
\usepackage[T1]{fontenc}
\usepackage[utf8]{inputenc}
\usepackage{microtype}
\usepackage{graphicx}
\usepackage{array}
\usepackage{booktabs}
\usepackage{multirow}
\usepackage{tabularx} 
\usepackage{threeparttable} 
\usepackage{rotating}  
\usepackage{amsmath}
\usepackage{graphicx}
\usepackage{subcaption}
\usepackage{graphicx}
\usepackage{array}
\usepackage{booktabs}
\usepackage{multirow}
\usepackage{tabularx}
\usepackage{threeparttable}
\usepackage{rotating}
\usepackage{amsmath}
\usepackage{xcolor}
\usepackage{subcaption}
\usepackage{colortbl}
\AtBeginDocument{%
  }

\copyrightyear{2026}
\acmYear{2026}
\setcopyright{cc}
\setcctype{by}
\acmConference[MM '26]{Proceedings of the 34th ACM International Conference on Multimedia}{November 10--14, 2026}{Rio de Janeiro, Brazil}
\acmBooktitle{Proceedings of the 34th ACM International Conference on Multimedia (MM '26), November 10--14, 2026, Rio de Janeiro, Brazil}
\acmDOI{10.1145/3767308.3836438}
\acmISBN{979-8-4007-2213-4/2026/11}

\begin{document}

\title{CLARA: Clip-Level Multimodal Alignment with VLM-Derived Rationales for Hateful Video Detection}

\author{Yuchen Zhang}
\orcid{0000-0001-8536-7175}
\affiliation{%
  \institution{Institute for Analytics and Data Science, University of Essex}
  \city{Colchester}
  \country{United Kingdom}
}
\email{yuchen.zhang@essex.ac.uk}

\author{Shuang Dai}
\orcid{0000-0001-7635-716X}
\affiliation{%
  \institution{Department of Engineering, University of Exeter}
  \city{Exeter}
  \country{United Kingdom}
  }
\email{s.dai@exeter.ac.uk}

\author{Zeyu Fu}
\orcid{0000-0002-2076-7597}
\correspondingauthor
\affiliation{%
  \institution{Department of Computer Science, University of Exeter}
  \city{Exeter}
  \country{United Kingdom}
  }
\email{z.fu@exeter.ac.uk}

\author{Yunfei Long}
\orcid{0000-0002-4407-578X}
\affiliation{%
 \institution{School of Electronic Engineering and Computer Science, Queen Mary University of London}
 \city{London}
 \country{United Kingdom}}
\email{qp241311@qmul.ac.uk}

\author{Ravi Shekhar}
\orcid{0000-0002-8798-641X}
\affiliation{%
  \institution{Institute for Analytics and Data Science, University of Essex}
  \city{Colchester}
  \country{United Kingdom}
}
\email{r.shekhar@essex.ac.uk}

\author{Haralambos Mouratidis}
\correspondingauthor
\orcid{0000-0002-2599-0712}
\affiliation{%
  \institution{Institute for Analytics and Data Science, University of Essex}
  \city{Colchester}
  \country{United Kingdom}
}
\email{h.mouratidis@essex.ac.uk}

\renewcommand{\shortauthors}{Yuchen Zhang et al.}

\begin{abstract}
Hateful video detection has become increasingly important with the rapid growth of video-centric social media platforms, given the serious risks that hate speech poses to both individual well-being and social cohesion. Compared with text or static multimodal content, hateful video detection remains underexplored and significantly more challenging, as hateful meaning often arises from complex interactions among multimodal cues, including speech, audio, and visual content. Moreover, such signals are often brief, implicit, and temporally dependent, making them difficult to capture using conventional video-level representations. In this work, we propose CLARA\footnote{The code for CLARA is available at \url{https://github.com/yuchenzhang-1/CLARA}.}, a clip-level multimodal framework for hateful video detection. Instead of treating a video as a single instance, CLARA models it as a sequence of fine-grained clips, enabling more precise capture of temporally localized hateful signals. We introduce a Mixture-of-Experts clip encoder for adaptive multimodal alignment, a local-global segment contrastive objective to jointly model short-term cues and long-range temporal dependencies, and VLM-derived rationales integrated via a gated Transformer to provide high-level semantic guidance. Extensive experiments on three hateful video datasets demonstrate that CLARA consistently outperforms state-of-the-art methods. Further ablation studies and parameter analyses validate the effectiveness of each component.

\textcolor{red}{\small \textbf{Disclaimer: This paper contains sensitive content that may be disturbing to some readers.}
}

\end{abstract}



\maketitle

\section{Introduction}

\begin{figure}[!t]
    \centering
    \includegraphics[width=0.7\linewidth]{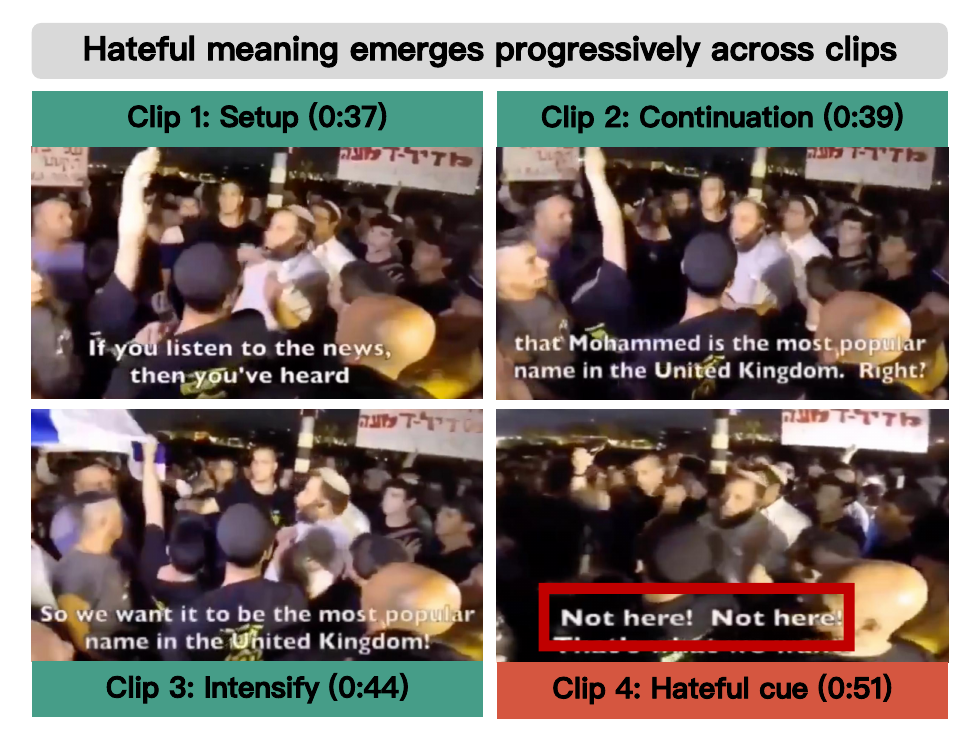}
    \caption{An example of hateful video. The hate speech is not fully revealed by single frame or utterance, but emerges through contextual buildup across multiple clips.}
    \label{fig:clip_motivation}
    \vspace{-1.5 em}
\end{figure}

Hate speech poses a serious threat to both individual well-being and social cohesion. It reinforces prejudice, normalizes discrimination, and intensifies hostility toward targeted groups, thereby causing psychological harm and deepening social division \cite{fortuna2018survey}. With the rapid growth of social media, particularly video-centric platforms such as TikTok, hateful content in video form is becoming increasingly prevalent. Accurate and efficient hate speech detection is therefore essential for enabling timely moderation and mitigating its broader societal harms.

Hate speech detection has been widely studied in text and, more recently, in multimodal settings such as memes and image-text posts \cite{yang2023hare, burbi2023mapping}. In contrast, hateful video detection remains relatively underexplored. This gap is increasingly important because videos present a substantially more challenging setting than text or static multimodal content. Hateful meaning in videos often arises from the interaction of spoken words, vocal tone, and visual scenes, with many cues being brief, context-sensitive, and deliberately indirect. Moreover, as illustrated in Figure~\ref{fig:clip_motivation}, hateful intent may not be fully revealed by single frame or utterance, but instead emerge progressively through contextual buildup across multiple clips of the whole video. A segment that appears neutral in isolation may reveal clear hateful intent only when interpreted in the context of preceding and subsequent content. Such temporal dependency makes hateful video understanding fundamentally more complex and challenging.

Existing hateful video detection methods still largely follow a coarse video-level modeling paradigm, where each modality is encoded over the entire video, reduced to a global representation, fused across modalities, and then used for final classification. Although recent advances have introduced stronger multimodal learning mechanisms, including cross-modal attention \cite{cespedes2025mm}, retrieval-augmented multimodal experts \cite{lang2025biting}, and vision-language-model (VLM) reasoning \cite{jing2025hvguard}, most existing approaches remain centered on video-level encoding and fusion. As a result, they remain limited in their ability to capture fine-grained and temporally evolving hateful signals.

In particular, three key challenges remain insufficiently addressed. 
First, the reliance on video-level representations prevents models from capturing fine-grained temporal units. This can be limiting, as hateful content in videos is often localized to short segments and may only become meaningful within specific contextual boundaries. Compressing the entire video into a single representation therefore risks diluting short but decisive hateful moments with long neutral content.
Second, existing methods typically rely on fixed multimodal fusion strategies, imposing a single fusion rule across the entire video. However, in hateful videos, the hateful evidence may shift across modalities over time, for example from spoken language to visual context \cite{cao2023multi}. Applying a uniform fusion mechanism thus fails to adapt to such dynamic modality contributions, limiting the model’s ability to capture diverse patterns of hateful expression.
Third, prior work rarely models the relationship between local cues and global context explicitly. This is a critical limitation because hateful meaning often emerges from the interaction between brief signals and broader discourse. Recent studies show that modeling both short-term and long-term temporal views can improve representation learning by preserving informative local structure while enforcing global temporal consistency \cite{wang2022long}, and similar findings have been reported in multimodal sequential learning \cite{jin2024rethinking}. These findings suggest that jointly modeling local and global dependencies is essential for accurately interpreting implicit and context-dependent hate.

To address these limitations, we propose \textbf{CLARA}, a \textbf{C}lip-\textbf{L}evel multimodal \textbf{A}lignment with VLM-derived \textbf{RA}tionales for hateful video detection. Instead of treating a video as a single instance, CLARA first segments it into video clips, preserving fine-grained temporal units. It then performs clip-level multimodal alignment using a Mixture-of-Experts (MoE) encoder, enabling input-dependent fusion across modalities. To bridge local cues and global context, we introduce a local-global segment contrastive objective that encourages the model to capture both short-term signals and long-range temporal dependencies. Finally, we incorporate VLM-derived rationales through a gated Transformer, which aggregates rationale and clip representations into a video-level representation, enabling the model to infer hateful meaning from interactions across clips under high-level semantic guidance for more effective hateful video detection.

Our contributions are summarized as follows:

\begin{itemize}
    \item We propose CLARA, a novel clip-level multimodal framework for hateful video detection that moves beyond coarse video-level modeling by explicitly representing videos as sequences of utterance-aligned clips, allowing decisive hateful cues to be preserved and modeled more effectively.

    \item CLARA combines a MoE-based clip encoder for adaptive multimodal alignment, a local-global contrastive objective to capture both short-term signals and long-range temporal dependencies, and VLM-derived rationales integrated via a gated Transformer to provide high-level semantic guidance, enabling more robust hateful video detection.

    \item Extensive experiments on three hateful video datasets demonstrate that CLARA consistently outperforms state-of-the-art baselines, achieving superior performance in hateful video detection, while ablation studies and parameter analyses further verify the effectiveness of its key design choices.
\end{itemize}

\section{Related Work}

Hate speech detection was initially studied mainly in text, and later extended to multimodal settings such as memes, image-text posts, and audio-based harmful content analysis \cite{hee2024recent}. In these settings, hateful meaning is often expressed jointly across modalities rather than through a single channel. For example, multimodal meme detection methods combine textual and visual reasoning for implicit hate analysis \cite{lin2023beneath, cao2023pro}, while audio-based approaches explore acoustic features for hate speech recognition \cite{yousefi2021audio,imbwaga2024automatic}. These developments are relevant to videos, but they do not directly address the key challenge of hateful video understanding, where speech, frames, and on-screen text unfold over time, and hateful meaning often arises from their interaction.

Currently, research on hateful video detection remains relatively limited. Early studies largely treated videos as text-bearing objects, relying on transcripts or metadata such as titles, descriptions, and comments, rather than modeling videos as multimodal temporal signals \cite{kandakatla2016identifying, alcantara2020offensive}. With the release of dedicated hateful video datasets such as HateMM \cite{das2023hatemm}, MultiHateClip \cite{wang2024multihateclip}, and DeHate \cite{zhang2025dehate}, subsequent work began to incorporate transcript, audio, and visual information for video-level hateful content prediction. A representative early multimodal baseline is CAMFusion \cite{zhang2024enhanced}, which extracts textual, visual, and acoustic features using BERT \cite{devlin2019bert}, ViT \cite{vit}, and MFCC, respectively, and fuses them through channel-wise and modality-wise operations before final classification.

More recent studies have explored stronger multimodal interaction mechanisms. For example, MM-HSD \cite{cespedes2025mm} enhances multimodal fusion by explicitly separating transcript and OCR text, extracting visual and audio features with ViT and Wav2Vec \cite{baevski2020wav2vec}, and applying cross-modal attention before final fusion. MoRE \cite{lang2025biting} introduces modality-specific experts, augments them with retrieved hateful and non-hateful neighbors through a joint multimodal retriever, and uses a sample-sensitive router to adapt modality contributions across short videos. HVGuard \cite{jing2025hvguard} instead adopts a reasoning-driven approach, where VLMs generate chain-of-thought-style rationales from video frames, transcripts, and audio emotion cues, and these rationale embeddings are then fused with modality representations through an MoE network. These studies show that stronger multimodal fusion, adaptive expert routing, retrieval augmentation, and explicit reasoning can substantially improve hateful video detection, especially for implicit and context-dependent hate.

Despite these advances, most existing methods still perform encoding and fusion at the whole video level. Each modality is compressed into a global representation before multimodal interaction and final prediction. Our work differs from this line by moving the basic modeling unit from the whole video to utterance-aligned clips, performing multimodal alignment at the clip level, and explicitly modeling the relationship between local hateful cues and broader discourse context before temporal aggregation.

\section{Proposed Method}

\begin{figure*}[!ht]
\centerline{\includegraphics[width=0.9\textwidth]{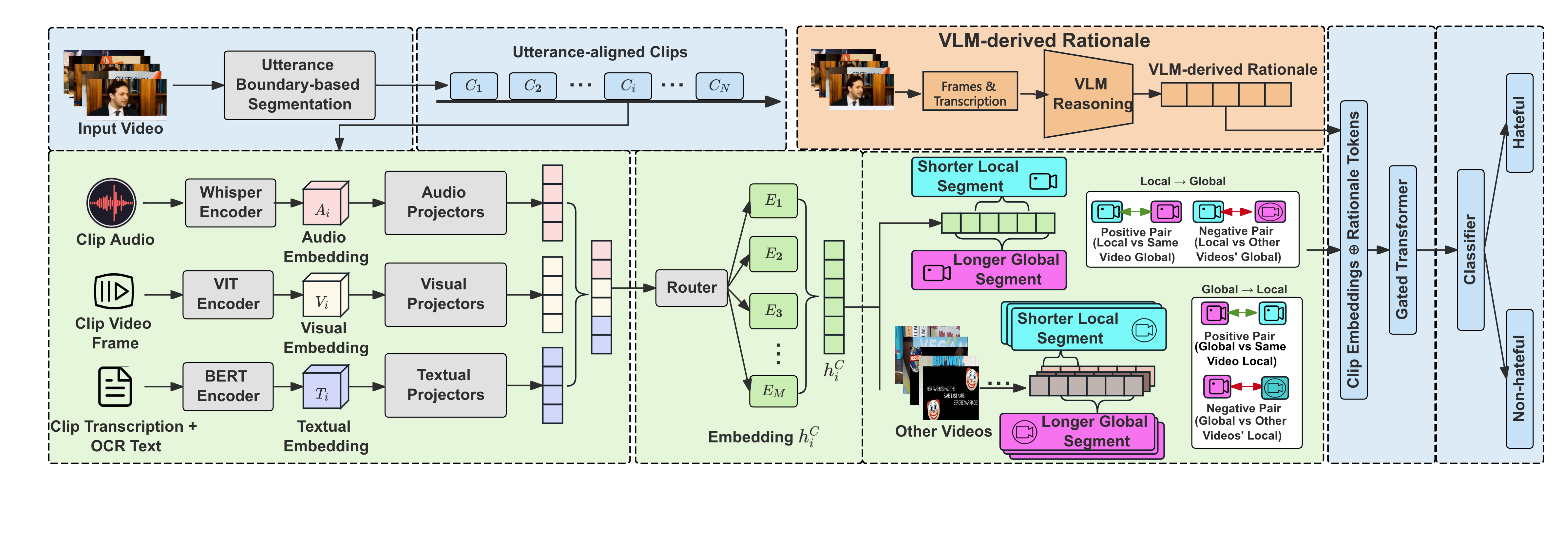}}
    \vspace{-1.5 em}
    \caption{Overall framework of CLARA. An input video is first segmented into utterance-aligned clips, from which multimodal features are extracted and fused into clip-level representations via a MoE encoder. Local-global segment contrastive learning is then applied to enhance temporal consistency across clips. Meanwhile, video-level rationales are generated by a VLM and integrated with the clip representations through a rationale-gated Transformer to produce the final video representation, which is finally used for hateful video classification. }
    \label{pic:framework}
\end{figure*}

Our goal is to detect hate speech in videos by capturing fine-grained clip-level multimodal interactions across text, audio, and visual signals, enhancing local-global temporal consistency, and leveraging VLM-generated video-level rationales as high-level semantic guidance, as illustrated in Figure~\ref{pic:framework}.
The proposed CLARA comprises six main components: (i) utterance-aligned clip segmentation, (ii) multimodal feature extraction and a MoE-based clip encoder, (iii) local-global segment contrastive learning, (iv) VLM-driven video-level rationale generation, (v) a rationale-gated Transformer that integrates the generated rationale with clip-level embeddings to produce a video-level representation, and (vi) a video-level hate speech classifer optimized with both supervised classification loss and a self-supervised contrastive objective.

\subsection{Video Clip Segmentation}

Videos exhibit highly heterogeneous temporal structures, where hateful cues may be diluted when modeling at the coarse video level.To better capture fine-grained multimodal interactions, we first segment each video into a sequence of clips.

Rather than fixed length segmentation, we segment videos based on utterance boundaries, as they provide explicit semantic units that align naturally with the underlying video content. Specifically, we employ Whisper-large-v3 \cite{radford2022whisper} to obtain transcriptions along with sentence-level timestamps, and use these timestamps to define clip boundaries. This ensures that each clip corresponds to a semantically coherent segment while remaining temporally aligned across text, audio, and visual modalities.

This segmentation process may introduce silent intervals between consecutive utterances. We then use the duration of each silent interval to determine whether it should be merged into an adjacent clip or retained as a standalone segment. Prior work suggests that pauses longer than 1 second are typically categorized as long pauses, and such pauses are more likely to mark boundaries between discourse units than to reflect brief within-utterance hesitations \cite{campione2002large}. We therefore use 1 second as the threshold, merging short pauses into adjacent clips to avoid unnecessary fragmentation while retaining longer silent spans as standalone segments.

\subsection{Multimodal Feature Extraction and MoE Clip Encoder}

Given a video $V_j = \{C_1,C_2, \dots, C_{N}\}$, where $N$ is the number of clips in video $V_j$, we first construct multimodal inputs for each clip $C_i \in V_j$, extract modality-specific features from text, audio, and visual streams, and then fuse them into a unified clip representation $h^C_i$ using a MoE encoder.

\paragraph{\textbf{Audio features.}}

For the audio modality, we feed the audio signal of $C_i$, denoted as $C_i^A$, into the encoder of Whisper-large-v3 \cite{radford2022whisper} to obtain the raw audio embedding $E_i^A$. 

\paragraph{\textbf{Visual features.}}
We define a fixed frame budget $S_{frame}$ for the entire video and allocate to each clip $C_i$ a clip-specific budget $s_i$ according to its duration. We then sample $s_i$  frames $ C_i^V = \{f^i_1, \dots, f^i_{s_i}\}$, and encode them using a pre-trained ViT \cite{vit} to obtain frame-level visual embedding $E_i^V$.

\paragraph{\textbf{Text features.}}
We use both transcription and OCR text as the textual input of each clip. Specifically, the transcription $C_i^{\text{trans}}$ and the OCR text $C_i^{\text{ocr}}$ extracted from the sampled frames $C_i^V$ using PaddleOCR \cite{cui2025paddleocr30technicalreport} are encoded using BERT \cite{devlin2019bert} to obtain $E_i^{\text{trans}}$ and $E_i^{\text{ocr}}$, respectively. The two embeddings are then concatenated to form the textual embedding $E_i^T = [E_i^{\text{trans}}, E_i^{\text{ocr}}]$.

\paragraph{\textbf{MoE-based clip encoder.}}

The extracted modality-specific embeddings are first mean-pooled and then projected into a shared latent space:
\begin{equation}
\tilde{E}_i^u
=
\phi_u\!\left(\mathrm{MeanPool}(E_i^u)\right),
\quad u \in \{A,V,T\},
\end{equation}
where $\phi_u(\cdot)$ is the corresponding modality-specific projector implemented as a two-layer MLP.

We then concatenate the modality-specific representations into a fused clip representation:
\begin{equation}
E_i^{Fused} = \mathrm{Concat}(\tilde{E}_i^A,\tilde{E}_i^V,\tilde{E}_i^T).
\end{equation}
Given the fused clip representation $E_i^{Fused}$, a gating network $g(\cdot)$ computes a routing distribution over $M$ experts:
\begin{equation}
p_i = \mathrm{softmax}(g(E_i^{Fused})), \quad p_i \in \mathbb{R}^M,
\end{equation}
where $g(\cdot)$ is implemented as a linear projection layer.

Based on $p_i$, we select the top-$K$ experts and compute their outputs as:
\begin{equation}
z_i^{(m)} = f_m(E_i^{Fused}), \quad m \in \mathrm{TopK}(p_i),
\end{equation}
where each expert $f_m(\cdot)$ is implemented as a two-layer MLP.

The final clip representation is obtained as a weighted combination of the selected expert outputs:
\begin{equation}
h_i^{C} = \sum_{m \in \mathrm{TopK}(p_i)} p_i^{(m)} z_i^{(m)}.
\end{equation}

To prevent routing collapse and encourage balanced expert utilization, we further introduce a load-balancing regularization term following prior MoE work \cite{shazeer2017outrageously, fedus2022switch}.  We define the importance of expert $m$ as the average routing probability assigned to it over a batch of size $B$:
\begin{equation}
\mathrm{Imp}(m) = \frac{1}{B} \sum_{i=1}^{B} p_i^{(m)}.
\end{equation}

We further define the load of expert $m$ as the fraction of top-$K$ assignments routed to it:
\begin{equation}
\mathrm{Load}(m)=\frac{1}{B}\sum_{i=1}^{B}\frac{1}{K}\,\mathbb{I}[m \in \mathrm{TopK}(p_i)],
\end{equation}
where $\mathbb{I}[\cdot]$ is the indicator function, which equals 1 if the condition holds and 0 otherwise. The load-balancing loss is then given by
\begin{equation}
\mathcal{L}_{\mathrm{LB}} = M \sum_{m=1}^{M} \mathrm{Imp}(m)\,\mathrm{Load}(m).
\end{equation}

\subsection{Local-Global Segment Contrastive Learning}

We introduce a local-global segment contrastive learning objective to encourage local segments to align with their corresponding global context while remaining distinguishable from segments of other videos, by enforcing consistency between short local segments and longer global segments within the same video and pushing apart segments from different videos.

Given a video $V_j = \{C_1, \dots, C_N\}$ with corresponding clip embeddings $\{h_1^C, \dots, h_N^C\}$, we construct two types of segment-level representations: (i) local segments consisting of consecutive short clips, and (ii) global segments covering a longer temporal span. 
Their lengths are determined as $s_\ell = \max(1, \lfloor r_\ell N \rfloor)$ and $s_g = \max(1, \lfloor r_g N \rfloor)$, where $r_\ell$ and $r_g$ are the local and global ratios and $r_g > r_\ell$.

Specifically, we sample the local segment and the global segment separately from the same video, using two independently selected valid start indices rather than forcing them to share the same temporal boundary or requiring one to contain the other. The resulting segments are
$\mathcal{S}_\ell = \{ h_{i_\ell}^C, \dots, h_{i_\ell+s_\ell-1}^C \}$ and $ \mathcal{S}_g = \{ h_{i_g}^C, \dots, h_{i_g+s_g-1}^C \}$,
where $i_\ell$ and $i_g$ are independently drawn valid start indices.

We aggregate each segment into a segment-level representation via mean pooling:
\begin{equation}
h_\ell = \mathrm{MeanPool}(\mathcal{S}_\ell), \quad
h_g = \mathrm{MeanPool}(\mathcal{S}_g).
\end{equation}

We treat the local and global segment representations from the same video as a positive pair, and use segment representations from other videos in the batch as negatives. Specifically, we adopt a bidirectional InfoNCE objective:
\begin{equation}
\mathcal{L}_{\text{CL}}=\frac{1}{2}\left(\mathcal{L}_{\ell \rightarrow g}+\mathcal{L}_{g \rightarrow \ell}\right),
\end{equation}
where
\begin{equation}
\mathcal{L}_{\ell \rightarrow g}
=
-\frac{1}{B}\sum_{i=1}^{B}
\log
\frac{\exp(\mathrm{sim}(h_\ell^{(i)}, h_g^{(i)})/\tau)}
{\sum_{j=1}^{B}\exp(\mathrm{sim}(h_\ell^{(i)}, h_g^{(j)})/\tau)},
\end{equation}
and
\begin{equation}
\mathcal{L}_{g \rightarrow \ell}
=
-\frac{1}{B}\sum_{i=1}^{B}
\log
\frac{\exp(\mathrm{sim}(h_g^{(i)}, h_\ell^{(i)})/\tau)}
{\sum_{j=1}^{B}\exp(\mathrm{sim}(h_g^{(i)}, h_\ell^{(j)})/\tau)}.
\end{equation}
Here, $\mathrm{sim}(\cdot,\cdot)$ denotes cosine similarity and $\tau$ is a temperature parameter.

\subsection{Gated Transformer for Video Encoding}

To model inter-clip temporal dependencies at the video level, we further introduce a rationale-gated Transformer encoder. This module incorporates video-level rationales generated by VLMs to guide the contextualization and aggregation of clip representations

\paragraph{\textbf{Video-level VLM rationale}}

For each video $V_j$, we generate a video-level rationale using Qwen3VL-8B-Instruct \cite{bai2025qwen3} conditioned on both visual and textual evidence. To obtain stable and interpretable outputs, we adopt a two-step prompting strategy.

First, we ask the VLM to perform an objective analysis of the video: (i) describe what is visible in the sampled frames, (ii) summarize the key messages conveyed by the provided text, (iii) provide an overall objective summary by combining the visual and textual evidence, and (iv) explain how the visuals and the text relate to each other. The inputs to this step include the video title, the transcription, and 20 sampled frames. The resulting output is a structured objective analysis, denoted as $OA_j$.

Then, we feed the same inputs as in the previous step, together with the objective analysis $OA_j$, into the VLM and ask it to perform hateful content verification: (i) determine whether the video is hateful or non-hateful, (ii) identify whether the hatefulness is explicit or implicit if hate is detected, and (iii) provide the main reasons supporting the decision, focusing on observable or stated evidence rather than speculation. Together, these two steps produce the final video-level rationale, denoted as $RA_j$, which is used used to obtain the rationale representation for subsequent video-level encoding.

\paragraph{Rationale-Gated Transformer.}

Given a video $V_j = \{C_1, \dots, C_N\}$ with corresponding clip representations $\{h_1^C, \dots, h_N^C\}$ and the video-level rationale $RA_j$, we model temporal dependencies using a rationale-gated Transformer encoder. Our design is motivated by the observation that standard Transformers perform fine-grained token interactions via self-attention, but do not explicitly regulate the relative contribution of different information sources before such interactions \cite{vaswani2017attention}. Inspired by feature-wise conditioning and conditional information modulation \cite{perez2018film, alayrac2022flamingo, li2023blip}, we introduce a lightweight source-level gating mechanism to apply coarse-grained modulation to the rationale and clip representations prior to Transformer encoding. Such coarse-to-fine control is particularly beneficial for hateful video detection, where the relative importance of clip level evidence and high level semantic reasoning can vary substantially across videos.

Specifically, for a video $V_j$, we first obtain its rationale representation $H_j^{RA}$ by encoding its rationale $RA_j$ using BERT to get its embedding  $E^{RA}_j$, and then projecting it into the same hidden space as the clip representations:
\begin{equation}
H_j^{RA} = \mathrm{Linear}(E^{RA}_j).
\end{equation}
Let $H_j^C = \{h_1^C, h_2^C, \dots, h_N^C\}$
denote the sequence of clip representations for video $V_j$.
We then compute two source-level gates, one for the rationale branch and the other for the clip branch, based on their pooled representations:
\begin{align}
g_j^{RA} = \sigma (\mathrm{Linear}(\mathrm{MeanPool}(H_j^{RA})), \\
g_j^{C} = \sigma (\mathrm{Linear}(\mathrm{MeanPool}(H_j^{C}))),
\end{align}
where $\sigma(\cdot)$ denotes the sigmoid function.

The two gates are used to modulate the rationale and clip representations before temporal encoding:
\begin{equation}
\hat{H}_j^{RA} = g_j^{RA}\, H_j^{RA}, \quad
\hat{h}_i^C = g_j^{C} h_i^C \quad (h_i^C \in H_j^C).
\end{equation}

We then concatenate the gated rationale and clip representations, and feed them into a Transformer encoder with positional encoding to obtain the final video representation of $V_j$:
\begin{equation}
H_j^{\text{final}} = \mathrm{Transformer}((\mathrm{PosEnc}([\hat{H}_j^{RA}, \hat{h}_1^C, \dots, \hat{h}_N^C])).
\end{equation}

\subsection{Training Objective}

Given the final video representation $H_j^{\text{final}}$, we perform video-level classification using an MLP classifier:
\begin{equation}
\hat{y}_j = \mathrm{softmax}(\mathrm{MLP}(H_j^{\text{final}})),
\end{equation}
where the MLP consists of two linear layers with a ReLU activation and dropout in between.

We optimize the model using a standard cross-entropy loss for supervised hateful video classification:
\begin{equation}
\mathcal{L}_{\text{CE}} = - \frac{1}{B} \sum_{j=1}^{B} y_j \log \hat{y}_j,
\end{equation}
where $y_j$ is the ground-truth label and $B$ is the batch size.

In addition, we have $\mathcal{L}_{\mathrm{LB}} $from MoE clip encoder and $\mathcal{L}_{\mathrm{CL}} $from local-global segment contrastive learning. The overall training objective is:
\begin{equation}
\mathcal{L} = (1- \lambda_{\text{CL}} )\mathcal{L}_{\text{CE}} + \lambda_{\text{CL}} \mathcal{L}_{\text{CL}} + \lambda_{\text{LB}} \mathcal{L}_{\text{LB}},
\end{equation}
where $\lambda_{\text{CL}}$ and $\lambda_{\text{LB}}$ control the contributions of the contrastive and load-balancing terms, respectively.

\section{Experiments}

\subsection{Experimental Setup}

\paragraph{\textbf{Datasets}}

\begin{table}[!htbp]
\vspace{-1 em}
\centering
\caption{Dataset Statistics. }
\vspace{-1 em}
\label{tab:dataset}
\resizebox{\linewidth}{!}{
\begin{tabular}{lcccc}
\toprule
\textbf{} &  \textbf{HateMM} & \textbf{MHC\_CN}  & \textbf{MHC\_EN}  & \textbf{DeHate} \\
\midrule
Hateful videos  &  427 & 298 & 296 & 2,120 \\
Non-hateful Videos  & 638 & 608 & 607 & 4,568 \\
Total videos &  1,065 & 906 & 903 & 6,688 \\
Platforms & BitChute & Bilibili  & YouTube & BitChute, TikTok \\
\bottomrule
\end{tabular}}
\end{table}

We conduct experiments on three publicly available hateful video datasets, namely HateMM \cite{das2023hatemm}, MultiHateClip \cite{wang2024multihateclip}, and DeHate \cite{zhang2025dehate}, as shown in Table~\ref{tab:dataset}. For MultiHateClip, we follow its language split into the Chinese subset (MHC\_CN) and the English subset (MHC\_EN) and adopt the binary classification setting from the original paper by grouping hate and offensive samples into one class and treating non-hate samples as the other class. 
These datasets differ in scale, language, content, and source platforms, covering BitChute, YouTube, Bilibili, and TikTok, and therefore provide a diverse benchmark for evaluating hateful video detection. 

\begin{table*}[!htbp]
\centering
\caption{Overall Performance in terms of Acc, M-F1, M-Pre and M-Rec. Best in \textbf{bold} and second-best is \underline{underlined}.}
\vspace{-1 em}
\label{tab:overall_results}
\resizebox{\textwidth}{!}{%
\begin{tabular}{lcccccccc}
\hline
\multirow{2}{*}{\textbf{Method}}
& \multicolumn{2}{c}{\textbf{HateMM}}
& \multicolumn{2}{c}{\textbf{MHC\_CN}}
& \multicolumn{2}{c}{\textbf{MHC\_EN}}
& \multicolumn{2}{c}{\textbf{DeHate}} \\
\cline{2-9}
& \textbf{Acc} & \textbf{M-F1} & \textbf{Acc} & \textbf{M-F1} & \textbf{Acc} & \textbf{M-F1} & \textbf{Acc} & \textbf{M-F1} \\
\hline
BERT
& 0.724 $\pm$ 0.028 & 0.709 $\pm$ 0.032
& 0.696 $\pm$ 0.006 & 0.635 $\pm$ 0.019
& 0.674 $\pm$ 0.010 & 0.417 $\pm$ 0.020
& 0.651 $\pm$ 0.006 & 0.522 $\pm$ 0.030 \\

Roberta
& 0.729 $\pm$ 0.014 & 0.718 $\pm$ 0.025
& 0.671 $\pm$ 0.003 & 0.402 $\pm$ 0.001
& 0.672 $\pm$ 0.002 & 0.402 $\pm$ 0.001
& 0.683 $\pm$ 0.000 & 0.406 $\pm$ 0.000 \\

MFCC
& 0.653 $\pm$ 0.048 & 0.598 $\pm$ 0.066
& 0.646 $\pm$ 0.014 & 0.511 $\pm$ 0.020
& 0.656 $\pm$ 0.018 & 0.463 $\pm$ 0.031
& 0.677 $\pm$ 0.003 & 0.441 $\pm$ 0.030 \\

Wav2vec
& 0.696 $\pm$ 0.013 & 0.678 $\pm$ 0.016
& 0.675 $\pm$ 0.003 & 0.403 $\pm$ 0.001
& 0.674 $\pm$ 0.008 & 0.403 $\pm$ 0.003
& 0.683 $\pm$ 0.001 & 0.407 $\pm$ 0.001 \\

ViT
& 0.700 $\pm$ 0.031 & 0.699 $\pm$ 0.031
& 0.701 $\pm$ 0.007 & 0.553 $\pm$ 0.024
& 0.678 $\pm$ 0.014 & 0.438 $\pm$ 0.026
& 0.685 $\pm$ 0.010 & 0.542 $\pm$ 0.018 \\

ViViT
& 0.718 $\pm$ 0.012 & 0.702 $\pm$ 0.020
& 0.673 $\pm$ 0.032 & 0.558 $\pm$ 0.030
& 0.688 $\pm$ 0.019 & 0.546 $\pm$ 0.037
& 0.687 $\pm$ 0.004 & 0.542 $\pm$ 0.022 \\ \midrule

LLaVA
& 0.688 $\pm$ 0.013 & 0.439 $\pm$ 0.015
& 0.669 $\pm$ 0.003 & 0.401 $\pm$ 0.001
& 0.673 $\pm$ 0.003 & 0.402 $\pm$ 0.001
& 0.687 $\pm$ 0.002 & 0.415 $\pm$ 0.007 \\

Qwen3VL
& 0.738 $\pm$ 0.035 & 0.727 $\pm$ 0.035
& 0.709 $\pm$ 0.041 & 0.597 $\pm$ 0.074
& 0.708 $\pm$ 0.014 & 0.588 $\pm$ 0.027
& 0.657 $\pm$ 0.013 & 0.595 $\pm$ 0.012 \\ \midrule

HateMM$^{\ast}$
& 0.771 $\pm$ 0.033 & 0.770 $\pm$ 0.033
& 0.739 $\pm$ 0.035 & 0.668 $\pm$ 0.046
& 0.678 $\pm$ 0.022 & 0.542 $\pm$ 0.030
& 0.695 $\pm$ 0.021 & 0.613 $\pm$ 0.016 \\

MHC$^{\ast}$
& 0.793 $\pm$ 0.023 & 0.782 $\pm$ 0.023
& 0.738 $\pm$ 0.020 & 0.669 $\pm$ 0.030
& 0.688 $\pm$ 0.031 & 0.600 $\pm$ 0.047
& 0.688 $\pm$ 0.017 & 0.622 $\pm$ 0.016 \\

DeHate$^{\ast}$
& 0.754 $\pm$ 0.037 & 0.736 $\pm$ 0.044
& 0.688 $\pm$ 0.009 & 0.595 $\pm$ 0.022
& 0.675 $\pm$ 0.034 & 0.554 $\pm$ 0.053
& 0.679 $\pm$ 0.015 & 0.590 $\pm$ 0.006 \\

MoRE
& 0.813 $\pm$ 0.030 & 0.803 $\pm$ 0.032
& 0.691 $\pm$ 0.018 & 0.539 $\pm$ 0.108
& 0.697 $\pm$ 0.017 & 0.584 $\pm$ 0.070
& \underline{0.711 $\pm$ 0.011} & 0.616 $\pm$ 0.027 \\

HVGuard
& 0.815 $\pm$ 0.026 & 0.813 $\pm$ 0.027
& \underline{0.754 $\pm$ 0.024} & \underline{0.707 $\pm$ 0.046}
& \underline{0.710 $\pm$ 0.030} & 0.632 $\pm$ 0.034
& 0.707 $\pm$ 0.003 & 0.586 $\pm$ 0.034 \\

MM-HSD
& \underline{0.839 $\pm$ 0.028} & \underline{0.833 $\pm$ 0.029}
& 0.683 $\pm$ 0.016 & 0.654 $\pm$ 0.019
& 0.688 $\pm$ 0.017 & \underline{0.653 $\pm$ 0.017}
& 0.689 $\pm$ 0.018 & \underline{0.626 $\pm$ 0.015} \\

CLARA
& \textbf{0.879 $\pm$ 0.022} & \textbf{0.872 $\pm$ 0.024}
& \textbf{0.779 $\pm$ 0.016} & \textbf{0.734 $\pm$ 0.026}
& \textbf{0.763 $\pm$ 0.035} & \textbf{0.727 $\pm$ 0.042}
& \textbf{0.735 $\pm$ 0.009} & \textbf{0.659 $\pm$ 0.012} \\

\midrule

\textbf{Method} & \textbf{M-Pre} & \textbf{M-Rec} & \textbf{M-Pre} & \textbf{M-Rec} & \textbf{M-Pre} & \textbf{M-Rec} & \textbf{M-Pre} & \textbf{M-Rec} \\
\hline
BERT
& 0.719 $\pm$ 0.030 & 0.705 $\pm$ 0.033
& 0.697 $\pm$ 0.014 & 0.629 $\pm$ 0.016
& 0.475 $\pm$ 0.196 & 0.503 $\pm$ 0.004
& 0.617 $\pm$ 0.012 & 0.543 $\pm$ 0.019 \\

Roberta
& 0.722 $\pm$ 0.024 & 0.703 $\pm$ 0.027
& 0.336 $\pm$ 0.001 & 0.500 $\pm$ 0.000
& 0.336 $\pm$ 0.001 & 0.500 $\pm$ 0.000
& 0.342 $\pm$ 0.000 & 0.500 $\pm$ 0.000 \\

MFCC
& 0.653 $\pm$ 0.085 & 0.608 $\pm$ 0.052
& 0.550 $\pm$ 0.027 & 0.528 $\pm$ 0.015
& 0.536 $\pm$ 0.046 & 0.512 $\pm$ 0.012
& 0.527 $\pm$ 0.036 & 0.508 $\pm$ 0.010 \\

Wav2vec
& 0.688 $\pm$ 0.016 & 0.676 $\pm$ 0.017
& 0.337 $\pm$ 0.002 & 0.500 $\pm$ 0.000
& 0.338 $\pm$ 0.003 & 0.498 $\pm$ 0.003
& 0.542 $\pm$ 0.245 & 0.500 $\pm$ 0.001 \\

ViT
& 0.701 $\pm$ 0.033 & 0.699 $\pm$ 0.031
& 0.669 $\pm$ 0.011 & 0.570 $\pm$ 0.014
& 0.644 $\pm$ 0.195 & 0.512 $\pm$ 0.010
& 0.610 $\pm$ 0.023 & 0.555 $\pm$ 0.012 \\

ViViT
& 0.711 $\pm$ 0.011 & 0.703 $\pm$ 0.022
& 0.619 $\pm$ 0.054 & 0.566 $\pm$ 0.024
& 0.634 $\pm$ 0.044 & 0.563 $\pm$ 0.026
& 0.615 $\pm$ 0.009 & 0.556 $\pm$ 0.013 \\ \midrule

LLaVA
& 0.678 $\pm$ 0.120 & 0.513 $\pm$ 0.009
& 0.335 $\pm$ 0.001 & 0.498 $\pm$ 0.002
& 0.337 $\pm$ 0.001 & 0.500 $\pm$ 0.000
& 0.527 $\pm$ 0.091 & 0.501 $\pm$ 0.003 \\

Qwen3VL
& 0.726 $\pm$ 0.035 & 0.736 $\pm$ 0.036
& 0.705 $\pm$ 0.088 & 0.605 $\pm$ 0.057
& \underline{0.729 $\pm$ 0.029} & 0.589 $\pm$ 0.020
& 0.598 $\pm$ 0.013 & 0.593 $\pm$ 0.011 \\ \midrule

HateMM$^{\ast}$
& 0.772 $\pm$ 0.035 & 0.771 $\pm$ 0.033
& 0.706 $\pm$ 0.051 & 0.659 $\pm$ 0.042
& 0.624 $\pm$ 0.053 & 0.557 $\pm$ 0.016
& 0.639 $\pm$ 0.029 & 0.608 $\pm$ 0.015 \\

MHC$^{\ast}$
& 0.788 $\pm$ 0.025 & 0.779 $\pm$ 0.021
& 0.704 $\pm$ 0.026 & 0.660 $\pm$ 0.027
& 0.630 $\pm$ 0.043 & 0.600 $\pm$ 0.038
& 0.631 $\pm$ 0.020 & 0.618 $\pm$ 0.015 \\

DeHate$^{\ast}$
& 0.748 $\pm$ 0.039 & 0.733 $\pm$ 0.045
& 0.635 $\pm$ 0.013 & 0.595 $\pm$ 0.017
& 0.605 $\pm$ 0.069 & 0.563 $\pm$ 0.043
& 0.614 $\pm$ 0.014 & 0.588 $\pm$ 0.006 \\

MoRE
& 0.806 $\pm$ 0.032 & 0.801 $\pm$ 0.034
& 0.588 $\pm$ 0.146 & 0.572 $\pm$ 0.065
& 0.653 $\pm$ 0.027 & 0.595 $\pm$ 0.052
& \underline{0.668 $\pm$ 0.027} & 0.614 $\pm$ 0.023 \\

HVGuard
& 0.811 $\pm$ 0.026 & 0.814 $\pm$ 0.028
& \underline{0.725 $\pm$ 0.031} & \underline{0.707 $\pm$ 0.056}
& 0.672 $\pm$ 0.062 & 0.628 $\pm$ 0.032
& 0.657 $\pm$ 0.006 & 0.590 $\pm$ 0.026 \\

MM-HSD
& \underline{0.835 $\pm$ 0.030} & \underline{0.833 $\pm$ 0.029}
& 0.652 $\pm$ 0.019 & 0.662 $\pm$ 0.023
& 0.652 $\pm$ 0.017 & \underline{0.658 $\pm$ 0.019}
& 0.654 $\pm$ 0.013 & \underline{0.626 $\pm$ 0.016} \\

CLARA
& \textbf{0.874 $\pm$ 0.022} & \textbf{0.872 $\pm$ 0.028}
& \textbf{0.760 $\pm$ 0.030} & \textbf{0.726 $\pm$ 0.032}
& \textbf{0.732 $\pm$ 0.041} & \textbf{0.725 $\pm$ 0.043}
& \textbf{0.697 $\pm$ 0.016} & \textbf{0.650 $\pm$ 0.011} \\
\hline
\end{tabular}%
}
\end{table*}

\paragraph{\textbf{Baselines}}

We compare CLARA against a comprehensive set of baselines spanning both unimodal and multimodal settings. For unimodal baselines, we include strong modality-specific encoders: BERT \cite{devlin2019bert} and RoBERTa \cite{liu2019roberta} for text, ViViT \cite{arnab2021vivit} and ViT \cite{vit} for video, and Wav2Vec \cite{baevski2020wav2vec} and Whisper \cite{radford2023robust} for audio. For multimodal baselines, we consider both general-purpose VLMs and task-specific hateful video detection methods. We evaluate LLaVA1.5-7B \cite{liu2023visual} and Qwen3VL-8B-Instruct \cite{bai2025qwen3} through direct prompting, using the same prompting procedure and input information as in rationale generation, and take the final VLM decision as the hateful or non-hateful prediction. For task-specific baselines, we include HateMM$^*$ \cite{das2023hatemm}, MHC$^*$ \cite{wang2024multihateclip}, and DeHate$^*$ \cite{zhang2025dehate}, which are the original benchmark baselines introduced with their respective datasets. These methods follow a similar pipeline in which modality-specific features are extracted separately, and then concatenated into a unified video representation for classification. We further include more recent multimodal methods. MoRE \cite{lang2025biting} incorporates retrieval-augmented contextual knowledge with modality-specific experts and sample-sensitive multimodal fusion. HVGuard \cite{jing2025hvguard} introduces VLM-based Chain-of-Thought reasoning for hate interpretation and uses a MoE network for multimodal fusion. MM-HSD \cite{cespedes2025mm} integrates video, audio, transcripts, and OCR text, and applies cross-modal attention to model inter-modal interactions.

\paragraph{\textbf{Implementation Details}}

For all baselines, unless otherwise specified, we follow the parameter settings reported in their original papers. For unimodal baselines, features extracted by each encoder are fed into the same classifier as in CLARA. For HVGuard, we implement the VLM-based reasoning using Qwen3VL instead of GPT-4o for fair comparison.

For CLARA, the total number of frames per video, $S_{\text{frame}}$, is set to 40. We use BERT-base Chinese\footnote{\url{https://huggingface.co/google-bert/bert-base-chinese}} for Chinese text, and BERT-base uncased\footnote{\url{https://huggingface.co/google-bert/bert-base-uncased}} for English text and VLM-derived rationales. The clip encoder adopts an MoE architecture with 8 experts and top-3 routing, and the weight of the load-balancing loss, $\lambda_{\text{LB}}$, is set to 0.01. For local-global contrastive learning, the local and global segment ratios, $r_\ell $and $r_g$, are set to 0.2 and 0.8, respectively, and the contrastive loss weight $\lambda_{\text{CL}}$ is set to 0.3. CLARA is trained with AdamW using a learning rate of 3e-5, weight decay of 1e-4, and a warmup ratio of 0.05, with a batch size of 32.

All experiments are conducted under 5-fold cross-validation with a train: validation: test split of 7:1:2. All models are trained up to 50 epochs with early stopping on two NVIDIA L40 GPUs.

\paragraph{\textbf{Evaluation Metrics}} 
We evaluate each model's performance with regard to accuracy (Acc), macro-precision (M-Pre), macro-recall (M-Rec) and macro-F1 (M-F1).

\subsection{Overall Results}

Table~\ref{tab:overall_results} presents the overall performance of all methods on all datasets. CLARA delivers the strongest performance consistently across all datasets and all four evaluation metrics, highlighting the effectiveness and robustness of our approach for hateful video detection.

Several clear trends can be observed. First, unimodal methods are consistently weaker than multimodal ones, indicating that hateful video detection cannot be reliably solved from a single modality alone. Second, VLMs such as LLaVA and Qwen3VL are stronger than most unimodal baselines, but still fall short of task-specific methods, indicating that hateful video detection requires more specialized modeling of fine-grained multimodal features.
Third,
among multimodal hateful video detection methods, MM-HSD, MoRE, and HVGuard are the strongest baselines, but CLARA consistently outperforms them across all datasets. Compared with the second-best method, CLARA achieves absolute improvements of 2.4\% to 5.3\% in Acc, 2.7\% to 7.4\% in M-F1, 0.3\% to 3.9\% in M-Pre, and 2.2\% to 9.7\% in M-Rec across all datasets. These consistent gains establish CLARA as a new state-of-the-art method for hateful video detection.

The superior performance of CLARA suggests that effective hateful video detection depends not only on leveraging multiple modalities, but also on when and at what temporal granularity they are aligned. Although methods such as HVGuard and MM-HSD also leverage adaptive fusion and reasoning-enhanced multimodal modeling, CLARA stands out by shifting multimodal interaction from the whole-video level to the utterance-aligned clip level. This design mitigates a key limitation of global video modeling, where short but decisive hateful cues can be diluted by long neutral segments. By aligning modalities before global compression, CLARA allows the model to focus on the clips where hateful meaning actually emerges.

Moreover, CLARA explicitly links local evidence with broader discourse structure. The local-global contrastive objective encourages short clip sequences to stay semantically consistent with longer contextual segments from the same video, while the gated Transformer models dependencies across clips as hateful meaning unfolds over time. The rationale signal further guides interpretation at the sequence level, rather than being introduced only after the video has already been compressed into a single representation. In summary, these design choices enable CLARA to more effectively align and contextualize fine-grained multimodal evidence, which is critical for hateful video detection.

\subsection{Ablation Study}

\begin{figure*}[!t]
    \centering
    \begin{subfigure}[b]{0.24\textwidth}
        \centering
        \includegraphics[width=\linewidth]{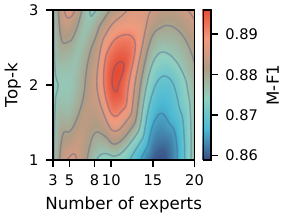}
        \caption{}
        \label{fig:sub1_moe}
    \end{subfigure}%
    \begin{subfigure}[b]{0.23\textwidth}
        \centering
        \includegraphics[width=\linewidth]{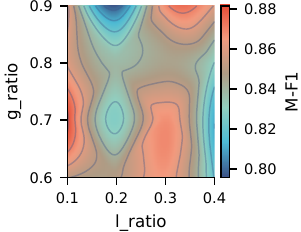}
        \caption{}
        \label{fig:sub2_lg}
    \end{subfigure}
       \begin{subfigure}[b]{0.22\textwidth}
        \centering
        \includegraphics[width=\linewidth]{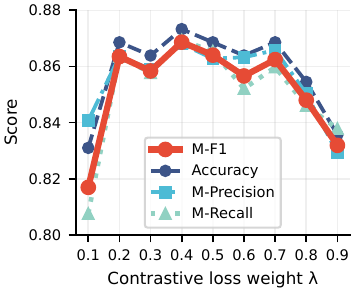}
        \caption{}
        \label{fig:sub3_cl}
    \end{subfigure}%
    \begin{subfigure}[b]{0.22\textwidth}
        \centering
        \includegraphics[width=\linewidth]{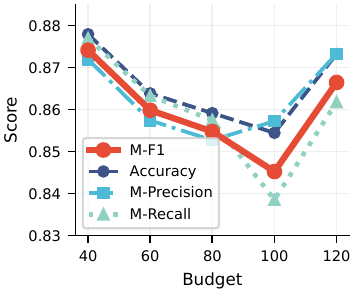}
        \caption{}
        \label{fig:sub4_budget}
    \end{subfigure}
    \caption{Parameter study of CLARA on HateMM. 
    (a) the effect of the MoE configuration, including the number of experts and top-k routing, (b) the effect of the local and global temporal ratios used in contrastive learning, (c) the effect of the contrastive learning weight $\lambda_{\text{CL}}$, and (d) the effect of the training budget.
    }
    \label{fig:three_figs}
\end{figure*}

\begin{table}[!htbp]
\centering
\vspace{-1 em}
\caption{Ablation Results in terms of Accuracy and M-F1.}
\vspace{-1 em}
\label{tab:ablation_results}
\resizebox{\linewidth}{!}{
\begin{tabular}{lcccccccc}
\hline
\multirow{2}{*}{Method}
& \multicolumn{2}{c}{HateMM}
& \multicolumn{2}{c}{MHC\_CN}
& \multicolumn{2}{c}{MHC\_EN}
& \multicolumn{2}{c}{DeHate} \\
\cline{2-9}
& Acc & M-F1 & Acc & M-F1 & Acc & M-F1 & Acc & M-F1 \\
\hline

w/o A
& 0.751 & 0.735
& 0.704 & 0.663
& 0.701 & 0.667
& 0.662 & 0.557 \\

w/o V
& 0.743 & 0.739
& 0.705 & 0.672
& 0.688 & 0.581
& 0.651 & 0.510 \\

w/o T
& 0.712 & 0.707
& 0.690 & 0.653
& 0.669 & 0.468
& 0.668 & 0.532 \\

\midrule

w/o MoE
& 0.789 & 0.780
& 0.738 & 0.679
& 0.735 & 0.669
& 0.708 & 0.637 \\

w/o CL
& 0.821 & 0.802
& 0.757 & 0.707
& 0.749 & 0.679
& 0.716 & 0.641 \\

w/o Rationale
& 0.815 & 0.803
& 0.708 & 0.645
& 0.727 & 0.719
& 0.683 & 0.506 \\

w/o Transformer
& 0.794 & 0.788
& 0.672 & 0.502
& 0.674 & 0.633
& 0.713 & 0.622 \\

\midrule

CLARA
& \textbf{0.879} & \textbf{0.872}
& \textbf{0.779} & \textbf{0.734}
& \textbf{0.763} & \textbf{0.727}
& \textbf{0.735} & \textbf{0.659} \\

\hline
\end{tabular}
}
\end{table}

To evaluate the contribution of each component in CLARA, we perform an ablation study by removing one component at a time from the full model. The variants `w/o A', `w/o V', and `w/o T' remove the audio, visual, and textual modalities, respectively. `w/o MoE' replaces the proposed MoE clip encoder with direct concatenation followed by an MLP projection. `w/o CL' removes the local-global contrastive objective. `w/o Rationale' excludes the VLM-derived rationale. `w/o Transformer' replaces the gated Transformer with mean pooling over clip representations. The results are reported in Table~\ref{tab:ablation_results}.

The results show that removing any component consistently reduces performance, indicating that the full model benefits from the interaction of all design choices. Dropping any modality leads to clear degradation across datasets, confirming that hateful video detection depends on complementary multimodal evidence rather than any single source alone. Among the modality variants, removing text causes the largest overall decline, while removing visual information also hurts performance noticeably. Similar trends are observed for the architectural components. Replacing the MoE with concatenation and MLP fusion weakens performance on all datasets, showing the value of adaptive clip-level fusion. Removing contrastive learning leads to consistent but smaller drops, suggesting that local-global alignment provides useful regularization beyond the base architecture. Removing rationale guidance further degrades performance, especially on M-F1, indicating that global semantic guidance helps improve balanced prediction. The performance drop becomes even more evident when the gated Transformer is replaced by mean pooling, particularly on MHC\_CN, which highlights the importance of explicit temporal modeling for capturing how hateful meaning unfolds across clips. Overall, the best results are achieved only when all components are included, demonstrating that multimodality features, clip-level alignment, contrastive learning, rationale guidance, and temporal aggregation contribute complementary benefits to hateful video detection.

\subsection{Parameter Analysis}

\paragraph{\textbf{MoE configuration.}}
We examine the impact of the MoE routing configuration by varying the number of experts in \{3, 5, 8, 10, 15, 20\} and the top-$k$ value in \{1, 2, 3\}. Figure~\ref{fig:sub1_moe} visualizes the resulting M-F1 scores as a heatmap, and warmer colors indicate better performance. 
The result reveals several clear patterns. First, configurations with a moderate number of experts, particularly around 8 to 10, consistently produce stronger results than very small or very large expert pools, with the peak performance achieved at 10 experts and top-2. This suggests that too few experts may limit the model ability to capture diverse multimodal patterns, whereas too many experts do not yield further gains and may instead weaken effective utilization. Second, the routing sparsity controlled by top-$k$ also plays an important role. Using top-1 tends to underperform in many cases, indicating that relying on only one expert can overly restrict collaboration. In contrast, larger routing choices such as top-3 do not bring further improvement over top-2, suggesting that activating too many experts may reduce specialization. Overall, the results show that a balanced MoE configuration, which provides sufficient expert diversity while preserving sparse and selective routing, offers the best trade-off for CLARA.

\paragraph{\textbf{Local-global ratio.}}
We further examine the impact of the local and global segment ratios used in contrastive learning, where the local ratio $r_\ell$ is varied over \{0.1, 0.2, 0.3, 0.4\} and the global ratio $r_g$ over \{0.6, 0.7, 0.8, 0.9\}, as shown in Figure~\ref{fig:sub2_lg}. The results do not show a monotonic trend with respect to either ratio. Instead, performance is sensitive to their combination, with the strongest performance concentrated around $r_\ell=0.2$, while $r_\ell=0.4$ leads to consistently weaker results under most global settings.
This pattern suggests that the local segment should remain sufficiently short to preserve fine-grained temporal cues. When the local ratio becomes too large, the distinction between local and global views becomes less clear, which weakens the contrastive signal. On the global side, increasing $r_g$ is generally beneficial, but only when paired with an appropriate local ratio. Overall, these results support the design choice of using distinct local and global temporal views, where a clear separation between the two is important for learning effective temporal representations.

\paragraph{\textbf{Contrastive learning weight.}}
We analyze the influence of the contrastive loss weight $\lambda_{CL}$ by varying it from 0.1 to 0.9, as shown in Figure~\ref{fig:sub3_cl}. The result shows that the performance improves as $\lambda$ increases from small values, reaches its best range at a moderate setting, and then gradually declines when $\lambda_{CL}$ becomes too large. In particular, the best overall result is obtained around $\lambda_{CL}=0.4$.
This trend indicates that CLARA benefits most from a balanced trade-off between the contrastive objective and the main classification objective. When $\lambda_{CL}$ is too small, such as 0.1, the contribution of contrastive learning is limited, preventing the model from fully exploiting temporal alignment. In contrast, when $\lambda_{CL}$ becomes too large, especially beyond 0.7, optimization places excessive emphasis on alignment learning at the expense of classification, which leads to degraded performance.

\paragraph{\textbf{Total Frame.}}
Finally, we evaluate CLARA under different total frame budgets ranging from 40 to 120. As illustrated in Figure~\ref{fig:sub4_budget}, the performance does not improve monotonically with increasing budget. CLARA achieves the best M-F1 when the budget is set to 40, while larger budgets does not necessarily lead to better performance.
This observation suggests that, at least on the HateMM dataset, a relatively small frame budget is sufficient for CLARA to capture the key multimodal cues for hateful video detection. Increasing the budget further does not provide additional benefits and may instead introduce redundant or less informative signals, which can negatively affect representation learning.

\subsection{MoE Routing Analysis}

\begin{figure}[t]
    \centering
    \begin{subfigure}[b]{0.25\textwidth}
        \centering
        \includegraphics[width=\linewidth]{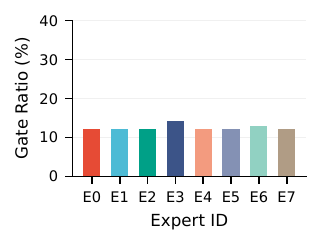}
        \caption{}
        \label{fig:routing_gate}
    \end{subfigure}%
    \begin{subfigure}[b]{0.25\textwidth}
        \centering
        \includegraphics[width=\linewidth]{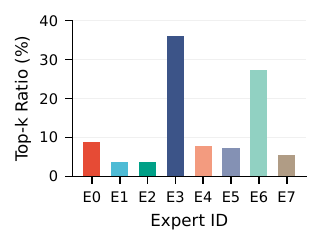}
        \caption{}
        \label{fig:routing_topk}
    \end{subfigure}
    \vspace{-2 em}
    \caption{Expert routing statistics of the MoE.
    : (a) the distribution measured by gate probability ratio. (b) the distribution of experts measured by top-k selection ratio.}
    \vspace{-1.5 em}
    \label{fig:routing_all}
\end{figure}

To better understand how the MoE module allocates computation across experts, we compare two routing statistics on the test set of HateMM dataset, namely \emph{Gate} and \emph{Top-k}. \emph{Gate} denotes the raw routing probabilities produced by the gating network before sparsification, reflecting the router's initial preference over all experts. In contrast, \emph{Top-k} denotes the effective routing probabilities after retaining only the top-k experts for each clip and renormalizing their scores, thus representing the actual expert usage during inference. Figure~\ref{fig:routing_all} visualizes the weighted expert ratio under these two distributions, where each bar indicates the proportion of total routing mass assigned to a given expert across all test clips.

As shown in Figure~\ref{fig:routing_gate}, the gate distribution is relatively balanced across experts, with most experts receiving a similar proportion of the total gating mass. This suggests that, before sparsification, the router does not exhibit an extremely peaked preference and still assigns non-negligible probability to multiple experts. However, after top-k selection, the effective routing distribution becomes much more concentrated. In particular, as shown in Figure~\ref{fig:routing_topk}, Expert~3 and Expert~6 receive substantially larger routing mass than the others, while the remaining experts are markedly suppressed. This pattern indicates that the top-k routing operation amplifies relatively small differences in the original gate scores, converting a moderately balanced soft distribution into a sparse and selective routing pattern.

Overall, these results suggest that CLARA does not rely on uniform expert usage. Instead, its MoE router first maintains a soft preference over multiple experts and then, through top-k sparsification, focuses computation on a small subset of experts that are consistently preferred. This behavior is consistent with the intended design of MoE, where sparse routing encourages selective expert specialization and more targeted processing at the clip level.

\section{Conclusion}

In this paper, we present \textbf{CLARA}, a clip-level multimodal framework for hateful video detection. Instead of compressing an entire video into a single representation, CLARA models fine-grained clips and their multimodal interactions over time. It integrates a MoE-based clip encoder for adaptive multimodal alignment, a local-global contrastive objective for cross-scale temporal consistency, and VLM-derived rationales through a gated Transformer for semantic guidance. Experiments on three hateful video datasets demonstrate consistent improvements over state-of-the-art methods, while ablation and sensitivity studies further verify the effectiveness of the proposed components.

\section{Acknowledgments}


This work has been partially funded by the European Union’s Horizon Europe research and innovation programme under the project ELOQUENCE (Grant Agreement No. 101135916) and the UKRI. Views and opinions expressed are, however, those of the author(s) only and do not necessarily reflect those of the European Union or the Research Executive Agency. Neither the European Union nor the granting authority can be held responsible for them.

\bibliographystyle{ACM-Reference-Format}
\bibliography{ref}

\newpage
\appendix 

\section{Clip Statistics}

Tables~\ref{tab:clip_count_statistics} and
\ref{tab:clip_duration_statistics} summarize the clip statistics after
utterance-aligned segmentation. HateMM and DeHate contain longer clip
sequences, with median values of 27 and 18 clips per video,
respectively, compared with 9 for both MHC\_CN and MHC\_EN. Clip
durations are nevertheless consistent across datasets, with median
values ranging from 2.0 to 2.4 seconds and most clips lasting
approximately 1.3 to 4.1 seconds. A small number of very short clips correspond to brief utterances, such as single-word expressions,
whereas the large maximum values in HateMM and DeHate reflect
long-tailed cases at the other end of the distribution..

\begin{table}[htbp]
\centering
\caption{Distribution of the number of clips per video.}
\label{tab:clip_count_statistics}
\resizebox{\columnwidth}{!}{
\begin{tabular}{cccccccc}
\toprule
\multirow{2}{*}{Dataset}
& \multirow{2}{*}{Videos}
& \multirow{2}{*}{Total Clips}
& \multicolumn{5}{c}{Clips per Video} \\
\cmidrule(lr){4-8}
& & & Mean & P25 & Median & P75 & Range \\
\midrule
DeHate  & 6,688 & 191,554 & 28.6 & 6 & 18 & 43 & 1--223 \\
HateMM  & 1,065 &  43,564 & 40.9 & 8 & 27 & 53 & 1--434 \\
MHC\_CN &   906 &  10,158 & 11.2 & 3 &  9 & 17 & 1--59  \\
MHC\_EN &   903 &   9,676 & 10.7 & 4 &  9 & 15 & 1--47  \\
\bottomrule
\end{tabular}
}
\end{table}

\begin{table}[htbp]
\centering
\caption{Distribution of clip durations.}
\label{tab:clip_duration_statistics}
\resizebox{\columnwidth}{!}{
\begin{tabular}{ccccccc}
\toprule
\multirow{2}{*}{Dataset}
& \multirow{2}{*}{Total Clips}
& \multicolumn{5}{c}{Clip Duration (s)} \\
\cmidrule(lr){3-7}
& & Mean & P25 & Median & P75 & Range \\
\midrule
DeHate  & 191,554 & 3.4 & 1.3 & 2.2 & 4.1 & 0.0$^{*}$--254.1 \\
HateMM  &  43,564 & 3.4 & 1.3 & 2.0 & 4.0 & 0.0$^{*}$--480.0 \\
MHC\_CN &  10,158 & 2.8 & 1.3 & 2.0 & 2.7 & 0.0$^{*}$--58.7 \\
MHC\_EN &   9,676 & 3.2 & 1.5 & 2.4 & 4.0 & 0.0$^{*}$--56.4 \\
\bottomrule
\multicolumn{7}{l}{\footnotesize $^{*}$Values below 0.05 s are rounded to 0.0.}
\end{tabular}
}
\end{table}

To bound the computational cost of video-level temporal modeling, we
set the maximum number of clips per video to 100 for CLARA and retain
the first 100 clips in temporal order when this limit is exceeded.
This threshold is well above the 75th percentile for every dataset and
therefore preserves the complete clip sequence for the vast majority
of videos. Specifically, it affects only 3.39\% of DeHate videos and
8.73\% of HateMM videos, while no MHC\_CN or MHC\_EN video exceeds
the limit. The maximum-length constraint primarily removes extreme
long-tail cases while providing a fixed upper bound on GPU memory and
the quadratic self-attention cost of the global video Transformer.

\section{Prompt Design for Video-level VLM Rationale Generation}

In this section, we provide additional details on how the video-level rationale $RA_j$ is generated in CLARA. Specifically, we describe the multimodal inputs provided to the VLM, the two-step prompting strategy, the exact prompts used in our experiments, and the structured generation procedure for obtaining the final rationale used in video-level encoding.

For each video $V_j$, we provide the VLM with two sources of information: (i) 20 uniformly sampled frames in temporal order, (ii) the video title and full transcription (If the video title is unavailable, we only provide the transcription). 
We use Qwen3-VL-8B-Instruct \cite{bai2025qwen3} as the VLM rationale generator with a two-step prompting strategy. In the first step, the VLM is asked to produce an objective analysis of the video. In the second step, the VLM performs hateful content verification conditioned on both the original multimodal inputs and the objective analysis in the first step. 

\subsection{Step 1: Objective Analysis}

In the first step, the VLM is prompted to describe the video content in a neutral and evidence-grounded way. The goal of this step is to obtain an objective analysis $OA_j$ that summarizes the visual content, textual content, the overall content summary, and their cross-modal relationship, without making any hateful content judgment.

For each video, the prompt specifies the textual inputs explicitly, whereas the visual evidence is supplied separately as image inputs to the VLM.
The exact prompt used in this step is shown in Table \ref{tab:prompt_step1}.

\begin{table}[!h]
\small
\centering
\setlength{\tabcolsep}{4pt}
\renewcommand{\arraystretch}{1.15}
\caption{Prompt used in Step 1 for objective video analysis.}
\begin{tabular}{p{0.97\linewidth}}
\hline
\specialrule{0.06em}{0em}{0em}

\rowcolor[HTML]{C0C0C0}
\textbf{STEP 1 PROMPT (Objective Analysis)} \\

\textcolor{gray}{\textit{\# Role}} \\
You are a professional video content verifier. Your task is to accurately summarize
what the video shows and what the accompanying text conveys in an objective and
neutral way. \\

\textcolor{gray}{\textit{\# Instructions}} \\
You may use: \\
(1) 20 frames sampled uniformly in temporal order. \\
(2) Video title (if available) and full transcription. \\

You need to: \\
(1) Describe what is visible in the frames. \\
(2) Summarize the key messages conveyed by the text. \\
(3) Provide an overall objective summary combining visual and textual evidence. \\
(4) Describe how the visuals and the text relate. \\
Return ONLY the following tagged format (no extra text before or after). Use the
tag names exactly: \\

\textcolor{gray}{\textit{\# Output Format}} \\
\textsc{VISUAL\_DESCRIPTION}: Neutral description of visible scenes, objects, actions, and
symbols. Avoid guessing details that are not supported by the frames. \\
\textsc{TEXTUAL\_DESCRIPTION}: Neutral summary of the title and transcription. Avoid guessing details that are not supported by the text. \\
\textsc{CONTENT\_SUMMARY}: Overall summary combining visual and textual evidence. \\
\textsc{CROSS\_MODAL\_RELATION}: One label from \{aligned, complementary, conflicting, unclear\}. \\
\textsc{CROSS\_MODAL\_EXPLANATION}: Briefly explain why that label fits, focusing on whether the visuals support, add context to, contradict, or do not clarify the text evidence. \\
\textsc{CONTEXT\_ELEMENTS}: List of key contextual elements that help interpret the content
(e.g., entities, groups, symbols). \\

\textcolor{gray}{\textit{\# Textual Input}} \\
VIDEO\_TITLE: \{\textit{video\_title}\} \\
TRANSCRIPTION: \{\textit{transcription}\} \\

\specialrule{0.06em}{0em}{0em}
\hline
\end{tabular}
\label{tab:prompt_step1}
\end{table}

\subsection{Step 2: Hateful Content Verification}

In the second step, the VLM is prompted to perform hateful content verification based on both the original multimodal inputs and the objective analysis $OA_j$ generated in the first step. This step asks the VLM to make the final judgment and provide evidence-based reasons. Compared with directly prompting for hateful or non-hateful classification in a single step, this design encourages the VLM to first establish grounded multimodal understanding before making a decision.
The exact prompt used in this step is shown in Table \ref{tab:prompt_step2}.

\begin{table}[]
\small
\centering
\setlength{\tabcolsep}{4pt}
\renewcommand{\arraystretch}{1.15}
\caption{Prompt used in Step 2 for hateful content verification.}
\begin{tabular}{p{0.97\linewidth}}
\hline
\specialrule{0.06em}{0em}{0em}

\rowcolor[HTML]{C0C0C0}
\textbf{STEP 2 PROMPT (Hateful Content Verification)} \\

\textcolor{gray}{\textit{\# Role}} \\
You are a professional content verifier conducting hateful content analysis. Your
task is to make a careful, evidence-based judgment using the information provided. \\

\textcolor{gray}{\textit{\# Instructions}} \\
You may use: \\
(1) 20 frames sampled uniformly in temporal order. \\
(2) video title and full transcription. \\
(3) The objective analysis of the video provided below. \\

You need to: \\
(1) Decide whether the video content is hateful or non-hateful. \\
(2) If hateful, indicate whether it is explicit or implicit. \\
(3) List the main reasons that support your decision, focusing on observable or
stated evidence rather than speculation. \\

Return ONLY the following tagged format (no extra text before or after). Use the
tag names exactly: \\

\textcolor{gray}{\textit{\# Output Format}} \\
\textsc{LABEL}: <hate | non-hate> \\
\textsc{EXPLICITNESS}: <explicit | implicit | na> \\
\textsc{CONFIDENCE}: <high|medium|low> \\
\textsc{REASONS}: <reason 1>, <reason 2>, ... \\
\textsc{NOTES}: <any brief clarifications or assumptions, if needed.> \\

\textcolor{gray}{\textit{\# Textual Information}} \\
VIDEO\_TITLE: \{\textit{video\_title}\} \\
TRANSCRIPTION: \{\textit{transcription}\} \\
OBJECTIVE ANALYSIS: \{\textit{objective\_analysis}\} \\

\specialrule{0.06em}{0em}{0em}
\hline
\end{tabular}
\label{tab:prompt_step2}
\end{table}

\subsection{Rationale Representation Generation}

\begin{table}[h]
\small
\centering
\setlength{\tabcolsep}{5pt}
\caption{Elements of textual rationale.}
\renewcommand{\arraystretch}{1.15}
\begin{tabular}{l p{0.7\linewidth}}
\hline
\textbf{Element} & \textbf{Description} \\
\hline

\textit{Content summary} & The \textsc{Content\_summary} from Step 1. \\

\textit{Visual description }& The \textsc{visual\_description} from Step 1. \\

\textit{Textual description} & The \textsc{textual\_description} from Step 1. \\

\textit{Cross-modal relation} & A sentence combining \textsc{cross\_modal\_relation} and \textsc{cross\_modal\_explanation} from Step 1 in the form ``\textit{The relation between textual and visual content is \{\textsc{cross\_modal\_relation}\} and \{\textsc{cross\_modal\_explanation}\}}''. If only one is available, we use it directly. \\

\textit{Contextual elements }& If \textsc{contextually\_elements} from Step 1 is a non-empty list, we convert it into a sentence of the form ``\textit{Contextually elements include: \(e_1, e_2, \dots, e_k\)}'', where \(e_i\) are the elements in \textsc{contextually\_elements}. Otherwise, this field is set to an empty string. \\

\textit{Final decision} & A sentence constructed from \textsc{Label}, \textsc{Explicitness}, and \textsc{Confidence} from Step 2. If the \textsc{Label} is non-hate, the sentence ``\textit{The video is considered non-hateful with \{\textsc{confidence}\} confidence.}'' If the label is hate, the sentence is ``\textit{The video is considered  hateful (\{\textsc{explicitness}\}) with \{\textsc{confidence}\} confidence.}'' \\

\textit{Reasons} & The \textsc{reasons} from Step 2. \\

\textit{Notes} & The \textsc{notes} from Step 2. \\

\hline
\end{tabular}
\label{tab:rationale_elements}
\end{table}

After these two steps, we construct the final rationale $RA_j$ as a structured list of strings:
$RA_j$ = [\textit{Content summary, Visual description, Textual description, Cross-modal relation, Contextual elements, Final decision, Reasons, Notes}]. 
Empty fields are preserved as empty strings so that all videos share the same rationale schema. Each element and its description in $RA_j$ are detailed in Table \ref{tab:rationale_elements}.

The resulting structured rationale $RA_j$ is then encoded using BERT \cite{devlin2019bert} to obtain its embedding $E^{RA}_j$:

\begin{equation}
    E^{RA}_j = \mathrm{BERT}(RA_j)
\end{equation}

\section{Statistical Significance Analysis}

To further verify that the performance gains of CLARA are not due to random variation across folds, we conduct additional pairwise t-tests at a 95\% confidence level (\(\alpha = 0.05\)). Specifically, we compare CLARA against three strong baselines, namely MoRE, HVGuard, and MM-HSD, on four evaluation metrics: Accuracy (Acc), Macro-F1 (M-F1), Macro-Precision (M-P), and Macro-Recall (M-R). The tests are conducted on the 5-fold cross-validation results for each dataset.

\begin{table}[h]
\centering
\small
\caption{Paired t-test p-values comparing CLARA with strong baselines across all datasets.}
\label{tab:ttest_all}
\resizebox{\linewidth}{!}{
\begin{tabular}{llccc}
\hline
\textbf{Dataset} & \textbf{Metric} & \textbf{CLARA--MoRE} & \textbf{CLARA--HVGuard} & \textbf{CLARA--MM-HSD} \\
\hline

\multirow{4}{*}{HateMM}
& Acc  & 4.00e-3 & 1.63e-2 & 6.50e-3 \\
& M-F1 & 6.10e-3 & 2.42e-2 & 7.50e-3 \\
& M-P  & 2.80e-3 & 2.42e-2 & 1.34e-2 \\
& M-R  & 9.40e-3 & 3.19e-2 & 1.17e-2 \\

\hline

\multirow{4}{*}{MHC\_CN}
& Acc  & 6.00e-4 & 6.50e-3 & 1.40e-3 \\
& M-F1 & 1.64e-2 & 3.00e-4 & 1.24e-2 \\
& M-P  & 3.94e-2 & 1.62e-2 & 4.10e-3 \\
& M-R  & 8.70e-3 & 1.00e-4 & 2.92e-2 \\

\hline

\multirow{4}{*}{MHC\_EN}
& Acc  & 5.10e-3 & 1.48e-2 & 9.00e-3 \\
& M-F1 & 7.00e-4 & 1.40e-3 & 2.19e-2 \\
& M-P  & 1.05e-2 & 7.60e-3 & 1.49e-2 \\
& M-R  & 1.00e-4 & 9.00e-4 & 3.77e-2 \\

\hline

\multirow{4}{*}{DeHate}
& Acc  & 1.77e-2 & 2.50e-3 & 8.40e-3 \\
& M-F1 & 1.56e-2 & 2.21e-2 & 8.20e-3 \\
& M-P  & 3.80e-3 & 6.80e-3 & 1.44e-2 \\
& M-R  & 1.55e-2 & 2.38e-2 & 2.47e-2 \\

\hline
\end{tabular}}
\end{table}

\begin{figure*}[!t]
    \centering
    \begin{subfigure}[b]{0.24\textwidth}
        \centering
        \includegraphics[width=\linewidth]{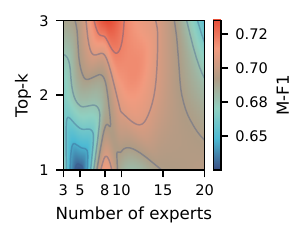}
        \caption{}
        \label{fig:MHC_CN_sub1_moe}
    \end{subfigure}%
    \begin{subfigure}[b]{0.23\textwidth}
        \centering
        \includegraphics[width=\linewidth]{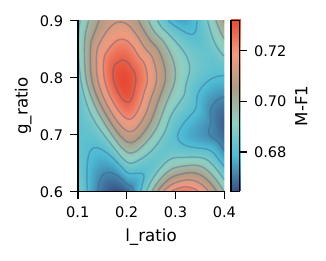}
        \caption{}
        \label{fig:MHC_CN_sub2_lg}
    \end{subfigure}
       \begin{subfigure}[b]{0.22\textwidth}
        \centering
        \includegraphics[width=\linewidth]{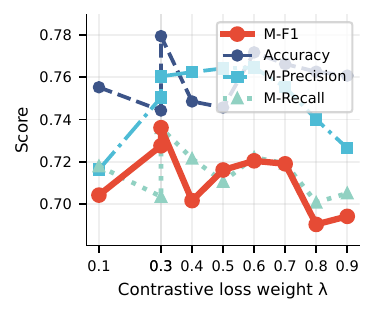}
        \caption{}
        \label{fig:MHC_CN_sub3_cl}
    \end{subfigure}%
    \begin{subfigure}[b]{0.22\textwidth}
        \centering
        \includegraphics[width=\linewidth]{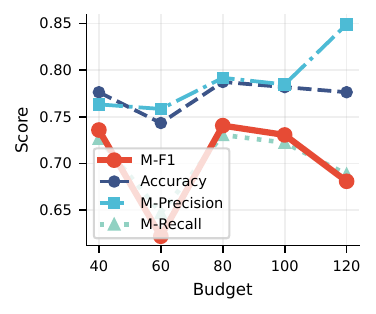}
        \caption{}
        \label{fig:MHC_CN_sub4_budget}
    \end{subfigure}
    \caption{Parameter study of CLARA on MHC\_CN. 
    (a) the effect of the MoE configuration, including the number of experts and top-k routing, (b) the effect of the local and global temporal ratios used in contrastive learning, (c) the effect of the contrastive learning weight $\lambda_{\text{CL}}$, and (d) the effect of the training budget.
    }
    \vspace{-1 em}
    \label{fig:MHC_CN_three_figs}
\end{figure*}

\begin{figure*}[!t]
    \centering
    \begin{subfigure}[b]{0.24\textwidth}
        \centering
        \includegraphics[width=\linewidth]{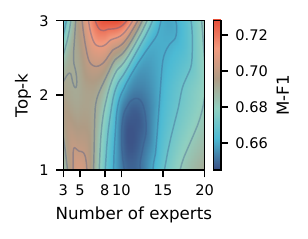}
        \caption{}
        \label{fig:MHC_EN_sub1_moe}
    \end{subfigure}%
    \begin{subfigure}[b]{0.23\textwidth}
        \centering
        \includegraphics[width=\linewidth]{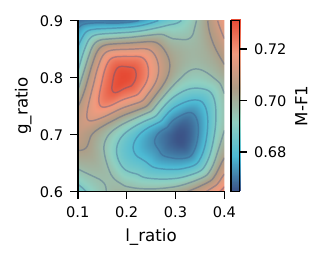}
        \caption{}
        \label{fig:MHC_EN_sub2_lg}
    \end{subfigure}
       \begin{subfigure}[b]{0.22\textwidth}
        \centering
        \includegraphics[width=\linewidth]{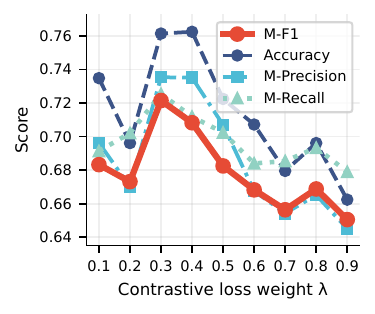}
        \caption{}
        \label{fig:MHC_EN_sub3_cl}
    \end{subfigure}%
    \begin{subfigure}[b]{0.22\textwidth}
        \centering
        \includegraphics[width=\linewidth]{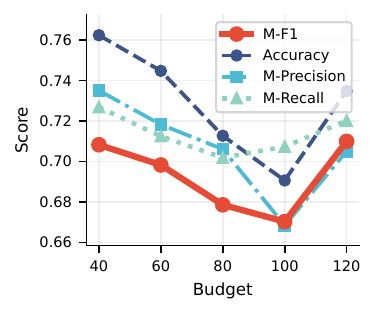}
        \caption{}
        \label{fig:MHC_EN_sub4_budget}
    \end{subfigure}
    \caption{Parameter study of CLARA on MHC\_EN.
    (a) the effect of the MoE configuration, including the number of experts and top-k routing, (b) the effect of the local and global temporal ratios used in contrastive learning, (c) the effect of the contrastive learning weight $\lambda_{\text{CL}}$, and (d) the effect of the training budget.}
    \vspace{-1 em}
    \label{fig:MHC_EN_three_figs}
\end{figure*}

\begin{figure*}[!t]
    \centering
    \begin{subfigure}[b]{0.24\textwidth}
        \centering
        \includegraphics[width=\linewidth]{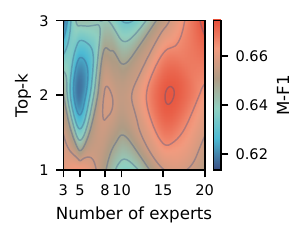}
        \caption{}
        \label{fig:DeHate_sub1_moe}
    \end{subfigure}%
    \begin{subfigure}[b]{0.23\textwidth}
        \centering
        \includegraphics[width=\linewidth]{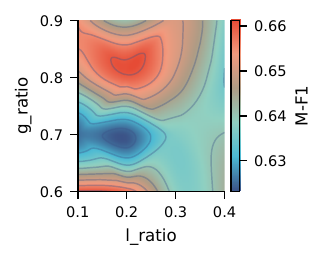}
        \caption{}
        \label{fig:DeHate_sub2_lg}
    \end{subfigure}
       \begin{subfigure}[b]{0.22\textwidth}
        \centering
        \includegraphics[width=\linewidth]{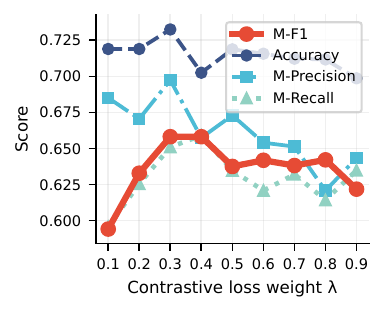}
        \caption{}
        \label{fig:DeHate_sub3_cl}
    \end{subfigure}%
    \begin{subfigure}[b]{0.22\textwidth}
        \centering
        \includegraphics[width=\linewidth]{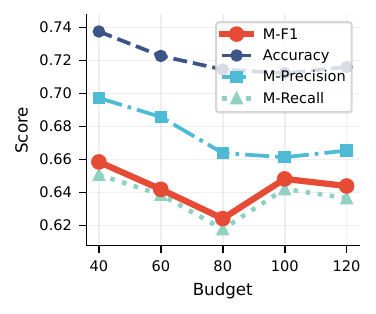}
        \caption{}
        \label{fig:DeHate_sub4_budget}
    \end{subfigure}
    \caption{Parameter study of CLARA on DeHate. 
    (a) the effect of the MoE configuration, including the number of experts and top-k routing, (b) the effect of the local and global temporal ratios used in contrastive learning, (c) the effect of the contrastive learning weight $\lambda_{\text{CL}}$, and (d) the effect of the training budget. }
    \vspace{-1 em}
    \label{fig:DeHate_three_figs}
\end{figure*}

As shown in Table~\ref{tab:ttest_all}, CLARA achieves statistically significant improvements over all three baselines across all datasets and all four evaluation metrics, with every \(p\)-value remaining below the 0.05 significance threshold. Beyond the overall significance, Table~\ref{tab:ttest_all} also reveals several dataset-specific patterns. 

First, on HateMM, all comparisons are statistically significant and the corresponding \(p\)-values remain consistently small across metrics and baselines, indicating stable improvements of CLARA over all competitors.
Second, MHC\_CN shows the largest variation across baselines and metrics. While all comparisons remain significant, the strength of significance differs substantially: some comparisons yield very small \(p\)-values (e.g., against HVGuard on M-F1 and M-R), whereas others are closer to the significance threshold (e.g., against MoRE on M-P).
Third, MHC\_EN presents a relatively consistent pattern, with most \(p\)-values remaining below 0.01 or close to that range, although a few comparisons, especially on M-R against MM-HSD, are closer to the threshold.
Finally, on DeHate, all comparisons also remain significant, with \(p\)-values distributed in a relatively narrow range. This suggests that the gains of CLARA on DeHate are statistically robust and more uniform across metrics, even though they are not as sharply separated as some of the strongest cases observed on HateMM or MHC\_CN.

Overall, these results further confirm that the improvements of CLARA are statistically reliable and consistently observed across different folds and datasets.

\section{Parameter Analysis on MultihateClip and DeHate}

\subsection{MoE Configuration}

We first examine the effect of the MoE configuration on MHC\_CN, MHC\_EN, and DeHate by varying the number of experts and the top-$k$ routing strategy. As shown in Figures~\ref{fig:MHC_CN_sub1_moe}, \ref{fig:MHC_EN_sub1_moe}, and \ref{fig:DeHate_sub1_moe}, a broadly consistent trend can be observed across all three datasets. CLARA performs best with a moderate MoE capacity rather than with overly small or overly large configurations. On MHC\_CN, the most favorable region appears when the expert pool is kept at a medium scale, between 8 and 15 experts, together with a more flexible routing strategy using top-2 or top-3 selection. MHC\_EN shows a similar pattern, with the strongest performance achieved when the number of experts remains at a moderate level, around 8 to 10. On DeHate, the optimal region shifts slightly toward a somewhat larger expert pool, while top-2 routing remains the most stable choice.

Overall, these results are generally consistent with the observation on HateMM. Increasing the number of experts from a very small setting improves the model ability to capture diverse multimodal patterns, suggesting that a certain degree of expert specialization is indeed beneficial. However, further enlarging the expert pool does not continuously improve performance and may even reduce it, likely because overly sparse or fragmented routing weakens training stability and makes specialization less effective. In the main experiments, we use 8 experts with top-3 routing as the default configuration, which provides sufficient expert capacity and allows flexible expert collaboration while keeping the routing complexity moderate. The results across the additional datasets further show that this configuration remains within a stable and competitive performance region.

\subsection{Local-global Ratio}

We further analyze the influence of the local and global temporal ratios used in contrastive learning. As illustrated in Figures~\ref{fig:MHC_CN_sub2_lg}, \ref{fig:MHC_EN_sub2_lg} and \ref{fig:DeHate_sub2_lg}, the three datasets exhibit a similar overall trend, since the best performance is obtained in an intermediate region of the ratio space. For both MHC\_CN and MHC\_EN, the highest M-F1 values are achieved when the local ratio is 0.2 and the global ratio is set in a relatively high range from 0.7 to 0.8. DeHate follows a similar pattern, with the most favorable region also appearing when the local ratio is kept small at 0.1 to 0.2 and the global context remains relatively large at 0.8 or above.

This result indicates that effective hateful video detection requires both short-span cues and broader contextual evidence, as observed on HateMM. The local branch helps preserve fine-grained signals that may only appear in a small number of clips, whereas the global branch provides the wider temporal context needed for interpretation. Although the exact optimum differs slightly across datasets, the general pattern remains stable. It can be observed that performance drops when the temporal design becomes extremely imbalanced, indicating that CLARA works best when local evidence and longer-range context are jointly modeled rather than overemphasizing only one temporal scale.

\subsection{Contrastive Learning Weight}

Next, we study the effect of the contrastive learning weight. As shown in Figures~\ref{fig:MHC_CN_sub3_cl}, \ref{fig:MHC_EN_sub3_cl}, and \ref{fig:DeHate_sub3_cl}, all three datasets consistently favor a moderate contrastive weight, with the most effective range lying between 0.3 and 0.4. On MHC\_CN, M-F1 reaches its highest value at 0.3. A similar pattern can be observed on MHC\_EN, where the best results also appear at 0.3, followed by a gradual decline as the contrastive objective becomes overly dominant. On DeHate, M-F1 increases clearly from smaller weights, reaches its peak at 0.3, and then decreases again as the weight continues to grow.

These results are consistent with the trend observed on HateMM. Introducing contrastive learning is clearly beneficial, as it helps align local and global temporal representations and improves the quality of multimodal feature learning. However, assigning too much weight to the contrastive objective can harm the final classification performance, because optimization becomes overly focused on representation alignment at the expense of the main discriminative objective. Therefore, a proper contrastive weight provides an effective trade-off between temporal alignment and hate classification.

\subsection{Total Frame}

Finally, we evaluate the effect of the total frame budget. As shown in Figures~\ref{fig:MHC_CN_sub4_budget}, \ref{fig:MHC_EN_sub4_budget} and \ref{fig:DeHate_sub4_budget}. While the specific strengths highlighted by these three datasets vary slightly, the overall conclusion remains consistent with that of HateMM. CLARA does not continue to benefit from an infinitely increasing frame budget. On MHC\_CN, performance improves from smaller budgets to a moderate budget, with the best M-F1 observed at 80 frames, while a further increase to 120 frames does not bring additional gains and instead leads to a noticeable drop in M-F1. On MHC\_EN, the strongest performance is already obtained with a relatively compact budget, and increasing the number of frames generally reduces the overall effectiveness. DeHate shows a similar trend, where smaller budgets are already sufficient to achieve the best or near-best M-F1, and larger budgets tend to introduce diminishing returns.

These results show a similar trend to that observed on HateMM. Once the key multimodal evidence has already been captured, adding more frames does not necessarily improve representation quality, and may instead introduce less informative content. This suggests that CLARA is able to identify useful hateful cues with a relatively compact temporal budget, while overly large inputs may dilute the most salient signals and make optimization less effective.

\section{MoE Routing Analysis on MultihateClip and DeHate}

\begin{figure}[!th]
    \centering
    \begin{subfigure}[b]{0.25\textwidth}
        \centering
        \includegraphics[width=\linewidth]{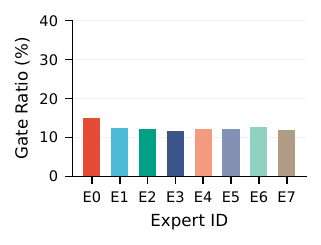}
        \caption{}
        \label{fig:MHC_CNrouting_gate}
    \end{subfigure}%
    \begin{subfigure}[b]{0.25\textwidth}
        \centering
        \includegraphics[width=\linewidth]{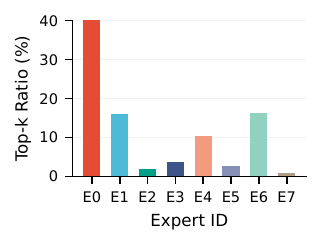}
        \caption{}
        \label{fig:MHC_CNrouting_topk}
    \end{subfigure}
    \vspace{-1 em}
    \caption{Expert routing statistics of the MoE on MHC\_CN
    : (a) the distribution measured by gate probability ratio. (b) the distribution of experts measured by top-k selection ratio.}
    \vspace{-1 em}
    \label{fig:MHC_CN_routing_all}
\end{figure}

\begin{figure}[!t]
    \centering
    \begin{subfigure}[b]{0.25\textwidth}
        \centering
        \includegraphics[width=\linewidth]{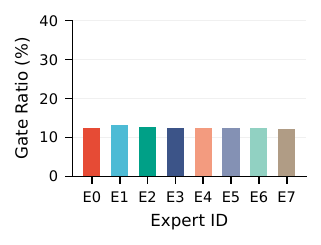}
        \caption{}
        \label{fig:MHC_EN_routing_gate}
    \end{subfigure}%
    \begin{subfigure}[b]{0.25\textwidth}
        \centering
        \includegraphics[width=\linewidth]{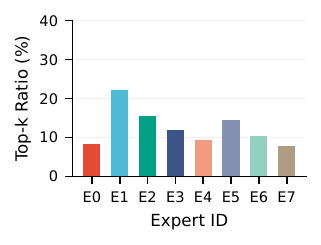}
        \caption{}
        \label{fig:MHC_EN_routing_topk}
    \end{subfigure}
    \vspace{-1 em}
    \caption{Expert routing statistics of the MoE on MHC\_EN
    : (a) the distribution measured by gate probability ratio. (b) the distribution of experts measured by top-k selection ratio.}
    \vspace{-1 em}
    \label{fig:MHC_EN_routing_all}
\end{figure}

\begin{figure}[!t]
    \centering
    \begin{subfigure}[b]{0.25\textwidth}
        \centering
        \includegraphics[width=\linewidth]{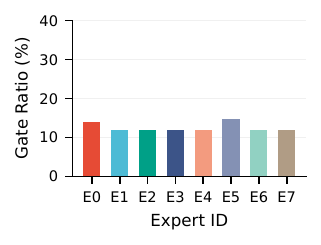}
        \caption{}
        \label{fig:DeHate_routing_gate}
    \end{subfigure}%
    \begin{subfigure}[b]{0.25\textwidth}
        \centering
        \includegraphics[width=\linewidth]{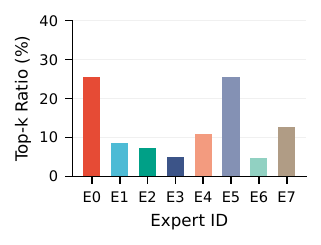}
        \caption{}
        \label{fig:DeHate_routing_topk}
    \end{subfigure}
    \vspace{-1 em}
    \caption{Expert routing statistics of the MoE on DeHate
    : (a) the distribution measured by gate probability ratio. (b) the distribution of experts measured by top-k selection ratio.}
    \vspace{-1 em}
    \label{fig:DeHate_routing_all}
\end{figure}

We further analyze the routing behavior of the MoE on MHC\_CN, MHC\_EN, and DeHate from two complementary perspectives: the gate probability ratio and the top-$k$ selection ratio. As shown in Figures~\ref{fig:MHC_CN_routing_all}--\ref{fig:DeHate_routing_all}, a consistent pattern can be observed across all three datasets. The gate probability distributions are relatively balanced across experts, while the top-$k$ selection distributions are much more uneven. This indicates that the gating network does not collapse to a dominant expert at the scoring stage. Instead, it maintains a broadly distributed preference over the expert pool, and the final sparse routing emerges mainly after top-$k$ selection. Such a discrepancy between soft gate scores and hard expert activation suggests that CLARA learns meaningful expert specialization rather than relying on uniformly shared routing.

Across the three datasets, the gate probability ratios remain relatively balanced across experts, indicating that all experts are retained as active candidates during routing. In contrast, the top-$k$ selection ratios are much more uneven, suggesting that the final routing decisions rely on a smaller subset of experts. MHC\_CN shows the strongest concentration pattern, where one expert dominates and only a few others are selected as secondary choices. MHC\_EN exhibits a more distributed routing structure, with several experts selected at comparable frequencies and no single expert clearly dominating. DeHate lies between these two cases, with two experts emerging as the main routing destinations while the remaining experts are activated much less frequently. These results suggest that although the gating network preserves broad routing flexibility, the final sparse routing still learns dataset-specific expert specialization patterns.

Overall, the routing statistics on all three datasets consistently support the design motivation of the MoE module in CLARA. The relatively balanced gate probability ratios show that the routing network preserves the potential utility of different experts, whereas the more skewed top-$k$ ratios confirm that the model can still make selective and sparse expert assignments when forming final representations.

\section{Impact of Model Components: Text Encoder and Rationale Generator}

\subsection{Impact of Text Encoder Variants}

We evaluate the impact of different text encoder choices on CLARA by comparing three variants: Qwen-based encoders \cite{qwen3embedding} with embedding dimensions of 1024 (qwen\_0.6) and 4096 (qwen\_8), as well as a BERT-based encoder. The results are reported in Table~\ref{tab:text_encoder_variants}.

Overall, CLARA demonstrates strong robustness across all text encoder variants, with consistently competitive performance on all datasets. Despite the substantial difference in embedding dimensionality between qwen\_0.6 (1024) and qwen\_8 (4096), the performance gap remains relatively small across all evaluation metrics. This indicates that the proposed framework is not sensitive to the specific representation scale of the textual features.
Moreover, while BERT achieves slightly better or comparable results on some datasets (e.g., HateMM), the improvements are marginal. In contrast, Qwen-based encoders, even with significantly different embedding sizes, maintain competitive performance across all datasets. This further highlights the flexibility of CLARA in accommodating heterogeneous text representations without requiring architecture-specific tuning.

These results suggest that CLARA is highly compatible with a wide range of text encoders, including those with different model architectures and embedding dimensionalities. Such compatibility is particularly important in multimodal settings, where textual representations may originate from diverse large language models.

\begin{table}[t]
\centering
\small
\caption{Impact of text encoder on CLARA across all datasets.}
\label{tab:text_encoder_variants}
\begin{tabular}{llcccc}
\hline
\textbf{Dataset} & \textbf{Text Encoder} & \textbf{Acc} & \textbf{M-F1} & \textbf{M-P} & \textbf{M-R} \\
\hline

\multirow{3}{*}{HateMM}
& qwen\_0.6 & 0.864 & 0.855 & 0.870 & 0.846 \\
& qwen\_8   & 0.878 & 0.871 & \textbf{0.882} & 0.864 \\
& bert      & \textbf{0.879} & \textbf{0.872} & 0.874 & \textbf{0.872} \\
\hline

\multirow{3}{*}{MHC\_CN}
& qwen\_0.6 & \textbf{0.780} & \textbf{0.736} & \textbf{0.767} & \textbf{0.730} \\
& qwen\_8   & 0.772 & 0.729 & 0.764 & 0.722 \\
& bert      & 0.779 & 0.734 & 0.760 & 0.726 \\
\hline

\multirow{3}{*}{MHC\_EN}
& qwen\_0.6 & 0.764 & 0.726 & 0.732 & 0.718 \\
& qwen\_8   & \textbf{0.778} & 0.721 & \textbf{0.733} & \textbf{0.725} \\
& bert      & 0.763 & \textbf{0.727} & 0.732 & 0.723 \\
\hline

\multirow{3}{*}{DeHate}
& qwen\_0.6 & \textbf{0.741} & \textbf{0.662} & \textbf{0.704} & \textbf{0.652} \\
& qwen\_8   & 0.730 & 0.637 & 0.692 & 0.629 \\
& bert      & 0.735 & 0.659 & 0.697 & 0.650 \\
\hline
\end{tabular}
\end{table}

\subsection{Impact of Rationale Generator Variants}

We further investigate the impact of different VLM-based rationale generators by comparing LLaVA1.5-7B \cite{liu2023visual} and Qwen3VL-8B-Instruct \cite{bai2025qwen3}, as shown in Table~\ref{tab:rationale_generator_variants}. 

It is worth noting that the two VLMs already exhibit a clear performance gap when directly applied to the hateful video detection task. As shown in Table~2 of the main manuscript, Qwen3VL consistently achieves markedly better performance than LLaVA across all datasets and all four evaluation metrics. This indicates that Qwen3VL provides stronger multimodal reasoning and alignment capabilities at the model level.
When integrated into CLARA as rationale generators, a similar trend is observed, where Qwen3VL-based rationales consistently lead to better performance across all datasets. This suggests that higher-quality multimodal reasoning from the VLM translates into more informative and discriminative rationale representations, which further benefit downstream classification.

Overall, these results show that rationale quality has a clear impact on the performance of CLARA, with stronger VLMs producing more informative rationales and leading to better downstream results. Nevertheless, even when using the weaker LLaVA, CLARA still remains competitive against strong baselines, suggesting that the framework can effectively leverage rationales of varying quality.

\begin{table}[t]
\centering
\small
\caption{Impact of rationale generator on CLARA across all datasets.}
\label{tab:rationale_generator_variants}
\begin{tabular}{llcccc}
\hline
\textbf{Dataset} & \textbf{VLM} & \textbf{Acc} & \textbf{M-F1} & \textbf{M-P} & \textbf{M-R} \\
\hline

\multirow{2}{*}{HateMM}
& LLaVA   & 0.8575 & 0.8443 & 0.8577 & 0.8379 \\
& Qwen3VL & \textbf{0.8790} & \textbf{0.8720} & \textbf{0.8740} & \textbf{0.8720} \\
\hline

\multirow{2}{*}{MHC\_CN}
& LLaVA   & 0.7084 & 0.6673 & 0.6715 & 0.6514 \\
& Qwen3VL & \textbf{0.7790} & \textbf{0.7340} & \textbf{0.7600} & \textbf{0.7260} \\
\hline

\multirow{2}{*}{MHC\_EN}
& LLaVA   & 0.7372 & 0.6898 & 0.7067 & 0.6859 \\
& Qwen3VL & \textbf{0.7630} & \textbf{0.7270} & \textbf{0.7320} & \textbf{0.7230} \\
\hline

\multirow{2}{*}{DeHate}
& LLaVA   & 0.7063 & 0.6103 & 0.6515 & 0.5979 \\
& Qwen3VL & \textbf{0.7350} & \textbf{0.6590} & \textbf{0.6970} & \textbf{0.6500} \\
\hline
\end{tabular}
\end{table}

\section{Role of Clip Segmentation Strategy}

We further evaluate the role of clip segmentation by replacing the utterance-aligned clips with fixed-length clips, where each video is split into consecutive 10-second clips. All other settings and hyperparameters are kept the same as in the full CLARA model. As shown in Table~\ref{tab:clip_segmentation_variants}, utterance-aligned segmentation consistently outperforms fixed-length segmentation across all datasets. The gap is largest on MHC\_CN, where M-F1 increases from 0.6547 to 0.7340, followed by HateMM, where it increases from 0.8285 to 0.8720. The gains are smaller but still consistent on MHC\_EN and DeHate, indicating that the benefit of utterance-level boundaries generalizes across datasets with different temporal characteristics.

These results show that semantically meaningful clip boundaries are important for CLARA. Fixed-length segmentation can split coherent utterances or mix unrelated content within the same clip, weakening clip-level multimodal alignment. In contrast, utterance-aligned segmentation better preserves semantic units and their corresponding visual, audio, transcript, and OCR evidence, which is especially useful when hateful meaning emerges from localized cues and their temporal context.

\begin{table}[!h]
\centering
\small
\caption{Impact of clip segmentation strategy.}
\label{tab:clip_segmentation_variants}
\begin{tabular}{llcccc}
\hline
\textbf{Dataset} & \textbf{Segmentation} & \textbf{Acc} & \textbf{M-F1} & \textbf{M-P} & \textbf{M-R} \\
\hline

\multirow{2}{*}{HateMM}
& Fix-length   & 0.8366 & 0.8285 & 0.8345 & 0.8288  \\
& Utterance-aligned & \textbf{0.8790} & \textbf{0.8720} & \textbf{0.8740} & \textbf{0.8720} \\
\hline

\multirow{2}{*}{MHC\_CN}
& Fix-length   & 0.7017 & 0.6547 & 0.7371 & 0.6452 \\
& Utterance-aligned & \textbf{0.7790} & \textbf{0.7340} & \textbf{0.7600} & \textbf{0.7260} \\
\hline

\multirow{2}{*}{MHC\_EN}
& Fix-length   & 0.7497 & 0.7073 & 0.7248 & 0.7041 \\
& Utterance-aligned & \textbf{0.7630} & \textbf{0.7270} & \textbf{0.7320} & \textbf{0.7230} \\
\hline

\multirow{2}{*}{DeHate}
& Fix-length  & 0.7110 & 0.6327 & 0.6567 & 0.6259  \\
& Utterance-aligned & \textbf{0.7350} & \textbf{0.6590} & \textbf{0.6970} & \textbf{0.6500} \\
\hline
\end{tabular}
\end{table}

\section{Computational Cost}

\begin{table}[!htbp]
\centering
\caption{Training and inference computational costs on HateMM.}
\label{tab:computation}
\resizebox{\columnwidth}{!}{
\begin{tabular}{lrrrrr}
\toprule
\multirow{2}{*}{Method}
& \multirow{2}{*}{Params (M)}
& \multicolumn{2}{c}{Training Cost}
& \multicolumn{2}{c}{Inference Cost} \\
\cmidrule(lr){3-4}
\cmidrule(lr){5-6}
&
& Time (s/epoch)
& Peak Mem. (MB)
& FLOPs (G/video)
& Latency (ms/video) \\
\midrule
MoRE    & 110.444 & 3.961 & 2,766.45 & 0.642 & 1.983 \\
MMHSD   &   4.626 & 0.329 &   352.29 & 0.008 & 1.052 \\
HVGuard &   3.312 & 0.205 &   133.96 & 0.007 & 0.552 \\
CLARA   &  44.950 & 3.515 & 2,765.14 & 2.033 & 3.720 \\
\bottomrule
\end{tabular}
}
\end{table}

We compare the computational cost of CLARA with representative
baselines on the HateMM dataset in terms of model parameters, training
time, peak GPU memory, FLOPs, and inference latency. As shown in
Table~\ref{tab:computation}, CLARA incurs a higher inference cost
than lightweight feature-level fusion methods, such as MMHSD and
HVGuard, due to its MoE-based clip encoder and global video Transformer.
Nevertheless, its training time and peak GPU memory remain comparable
to those of MoRE, while requiring substantially fewer parameters.
Considering the superior detection performance achieved by CLARA on
HateMM, these results demonstrate a favorable trade-off between
predictive performance and computational cost.

\end{document}